\documentclass[11pt]{article}

\usepackage{fullpage}

\usepackage[section]{placeins}

\usepackage[utf8]{inputenc} 
\usepackage[T1]{fontenc}    
\usepackage{url}            
\usepackage{booktabs}       
\usepackage{amsfonts}       
\usepackage{nicefrac}       
\usepackage{microtype}      
\usepackage{xcolor}         
\usepackage{subcaption}
\usepackage{graphicx}

\newlength{\twofigwidth}
\newlength{\twocapwidth}
\newlength{\threefigwidth}
\newlength{\threecapwidth}

\newcommand{\vecrm}{\mathrm{vec}}

\newcommand{\qed}{\hfill\rule{1ex}{1.5ex}}

\title{\bf Properties of the After Kernel}

\author{%
  Philip M. Long \\
  Google \\
  \texttt{plong@google.com} \\
}

\date{}

\begin{document}

\maketitle

\begin{abstract}
The Neural Tangent Kernel (NTK) is the wide-network limit of a kernel
defined using neural networks at initialization, whose embedding is
the gradient of the output of the network with respect to its
parameters.  We study the ``after kernel'', which is defined using the
same embedding, except after training, for neural networks with
standard architectures, on binary classification problems extracted
from MNIST and CIFAR-10, trained using SGD in a standard way.  
For some dataset-architecture pairs,
after a few epochs of neural network training,
a hard-margin SVM using the network's
after kernel is much more accurate than when the network's initial kernel
is used.
For networks with an architecture similar to VGG,
the after kernel is more ``global'', in the sense that it is less
invariant to transformations of input images that disrupt the global
structure of the image while leaving the local statistics largely intact.
For fully connected networks, the after kernel is less global
in this sense.
The after kernel tends to
be more invariant to small shifts, rotations and zooms;  data
augmentation does not improve these invariances.  The (finite
approximation to the) conjugate kernel, obtained using the last layer
of hidden nodes, sometimes, but not always, provides a good
approximation to the NTK and the after kernel.  

Training a network with a larger learning rate (while holding the training error constant)
produces a better kernel, as measured by the test error of
a hard-margin SVM.  The after kernels of networks trained with larger learning
rates tend to be more global, and more invariant to small shifts, rotations and zooms.
\end{abstract}

\section{Introduction}
\label{s:intro}

The neural tangent kernel (NTK)
\cite{DBLP:conf/nips/JacotHG18,DBLP:conf/iclr/DuZPS19,DBLP:conf/nips/ChizatOB19,DBLP:conf/nips/LiL18a,DBLP:conf/nips/LeeXSBNSP19} 
has generated intense interest in recent years
(see
\cite{DBLP:conf/iclr/DuZPS19,du2019gradient,DBLP:conf/nips/CaoG19a,allen2019convergence,DBLP:conf/nips/ZouG19,CCZG21,DBLP:conf/iclr/JiT20,DBLP:conf/icml/AroraDHLW19,CLB21}).
Its definition makes reference to a neural network at initialization.
If $\theta_0$ are the original parameters of the network, and
$f_{\theta_0}$ is the function computed by the network then,
\[
K_{\mathrm{FNKT}}(x, x') = \phi(x) \cdot \phi(x'), 
\]
where
$\phi(x)$ is the Jacobian of the function mapping $\theta$
to $f_{\theta}(x)$, evaluated at $\theta_0$.  
The NTK is a limit of $K_{\mathrm{FNTK}}$ (the ``F'' stands for
``finite'') as the width of a network goes to infinity
\cite{yang2019scaling,DBLP:conf/nips/AroraDH0SW19}.
Standard proofs using the NTK exploit the fact that,
under conditions on the width, initialization, and learning rate,
a first-order Taylor approximation of the objective function around
the initial solution remains accurate throughout training.  
That is, the algorithm is learning a linear classifier over
features that were determined at initialization.

A kernel analogous to the FNTK could also be defined after training.
In this paper, we call this
the {\em after kernel}.
The after kernel was previously studied by
Fort, et al \cite{DBLP:conf/nips/FortDPK0G20}, by evaluating
algorithms that replace the objective function of neural network
training with its first-order
Taylor expansion, after different amounts of training, effectively
training a model using the after kernel.  They showed that, after a few
epochs of training, the after kernel is ``better'' than the
NTK, and that additional epochs of training continue to improve it, but
more gradually.  
Baratin, et al \cite{baratin2021implicit}
showed experimentally that, for various
architectures on CIFAR10 and MNIST, much of the energy of the spectrum of the
after kernel is concentrated in a few eigenvalues, and that
the alignment of the kernel with 
the labels increases.

In this paper, we further study the after kernel.
We work with two binary classification problems defined
using MNIST (3-vs-8 and 5-vs-6), and one using CIFAR10 (cats-vs-dogs).
We use two architectures, one inspired by VGG \cite{SZ15}, along with
a depth-four, fully connected network.  We chose VGG because it is
relative simple and canonical, but includes several elements
that are part of the standard practice of neural networks, including
the use of convolutional layers, max pooling, and analysis of the
image on multiple scales.  We experimented with fully connected
networks because they are, in a sense, the most basic and canonical
neural networks, and, as such, are often the subject of theoretical
analysis.  We were especially interested in the extent to which
phenomena observed in fully connected networks could also be found in
convolutional networks.
Many of our experiments with
the VGG-like network use a network with approximately 100K parameters.
We also experiment with a larger network, which we call the
mega-VGG-like architecture, because it has roughly one million
parameters.  To obtain instances of the after kernel, we train
neural networks with these architectures using Keras
\cite{chollet2015} with the default hyper-parameters.  The details are
in Section~\ref{s:prelim}, and code is available 
online
\cite{Lon21akcode}.
One noteworthy point is that all results were averaged over ten runs,
where each run includes a random initialization of the network and
random permutations of the training data.

We begin by 
conducting an experiment along the lines of \cite{DBLP:conf/nips/FortDPK0G20},
comparing the after kernel with the FNTK.
To evaluate the quality of a 
kernel $K$, we evaluate the test-set error of 
the application of Algorithm 1 from \cite{ji2021fast}
to compute the hard-margin SVM for $K$.
Using the hard-margin SVM to evaluate the kernel is motivated in part
by \cite{LL20}, who, building on the work of
\cite{DBLP:conf/iclr/SoudryHNS18,ji2018gradient,DBLP:conf/colt/JiT19,DBLP:conf/icml/NacsonGLSS19,DBLP:conf/nips/WeiLLM19},
showed that, under certain conditions, training neural networks to
convergence produces a model equivalent to a hard-margin SVM applied
with the after kernel.  (\cite{ji2021fast} showed that their maximum
margin algorithm converges much faster than SGD on the softmax loss,
which was used to evaluate the kernel in the experiments in
\cite{DBLP:conf/nips/FortDPK0G20}.)  We show that, for fully connected
networks on the MNIST problems, and VGG-like networks on the
CIFAR10 cats-vs-dogs problem, the after kernel is much better than the
FNTK in this sense.

Further, as seen in \cite{DBLP:conf/nips/FortDPK0G20}
under related conditions, we see that,
even for the problems where the after kernel improves a lot on the
FNTK, most of the improvement is seen after a few epochs.  The test
error of the SVM trained on the after kernel is, for quite a while, a
lot better than the test error of the neural network.  Consistent with
the theory in \cite{LL20}, however, the neural network finally catches
up.  Informally, for much of training,
for the most part, the deep-learning algorithm is effectively learning 
a linear model over features that were largely determined much earlier.

Next, we address the question of how ``global'' the kernels
are.  To measure this, 
following \cite{DBLP:conf/nips/NeyshaburSZ20},
we create artificial images in
which the top left quadrant is swapped with the bottom right
quadrant.  Such swapping leaves local statistics largely
intact, but disrupts the global structure of the image.
To measure this invariance, we compute the average cosine
similarity between embeddings computed using the unperturbed
images with embeddings computed using the perturbed
images.  For the VGG-like networks, the after kernel is
substantially more global than the FNTK.  For fully connected
networks, however, this was not seen.  The average similarity
shot up after one epoch of training, after which it gradually
decreased, settling down at a larger value than what was seen in
the FNTK.

Next, we evaluate how invariant kernels are to slight
perturbations of the images that should not affect their
classifications.   We evaluate three perturbations:
shifts by one pixel, rotations by small angles, and
slight zooms (obtained by cropping out a layer of pixels
from the outside of the original image, and then resizing
the cropped image to the original size).  In most cases,
the after kernel is more invariant to these perturbations.

The conjugate kernel \cite{DBLP:conf/nips/DanielyFS16} can be defined
as the wide-network limit of a kernel whose embedding is obtained
from a layered neural network by using the activations from the layer
closest to the output layer.  
The components of
the embedding of the conjugate kernel are also present in the FNTK.
Accordingly, we will refer to an FNTK and its {\em conjugate
  projection}.  The conjugate kernel was originally defined and
analyzed using the network at initialization, but there is a natural
counterpart after training, which we will call the conjugate
projection of the after kernel.  We evaluate the similarity between
the FNTK and its conjugate projection using kernel alignment
\cite{NIPS2001_1f71e393}.  For VGG-like networks, the conjugate
projection, whose embedding has many fewer features, and which is much
easier to compute, provides a good approximation to the FNTK, and the
after kernel is accurately approximated by its conjugate projection.
For fully connected networks on the MNIST problems, however, the conjugate
projection does not approximate the after kernel.

When its conjugate projection approximates the after kernel,
as might be expected, the accuracy of SVM models using the
respective kernels are similar.  For VGG networks on
the MNIST 5-vs-6 problem, however, the after kernel leads
to better accuracy.
On the CIFAR10
cats-vs-dogs data, where the after kernel and its conjugate projection
are different, the after kernel leads to better accuracy for a while,
but its conjugate projection catches up.

Next, we evaluate the effect of data augmentation on the learned
kernel.  We add two kinds of augmentation, random shifts and
random rotations.  
Perhaps surprisingly, adding these
augmentations tends not to improve the invariance of the after kernel
to shifts, rotations and zooms, despite the fact that augmentations
improve the test error, as seen in previous work.

In all of the above experiments, we used Keras' default learning rate
when training the networks.  
NTK proofs typically require a small learning rate, whereas models trained
with a larger learning rate tend to generalize better \cite{mandt2017stochastic,DBLP:conf/iclr/SmithL18}.
This is usually attributed to the tendency of training with a large learning rate to
land in ``wide minima''.  However, the fact that small learning rates are required
for staying in the NTK regime raises the question of whether another reason that
large learning rates help is that they facilitate feature learning.  Here, we evaluate
this hypothesis using natural data with natural architectures.  For each learning rate,
we train a network until the training error reaches 0.01, and compare the after kernels.
We find that the kernel of the networks trained with larger learning rates are better,
in the sense that using them with SVM produces better test error.  Kernels produced using 
the larger learning rates are also more global, as measured using invariance to
swaps between the upper left and lower right quadrants.  The large-learning rate
kernels are also more invariant to small translations, rotations and zooms.  
Finally, while hard-margin classifiers trained with the after kernel are generally
more accurate than when the conjugate kernel is used, the gap is smallest when the
network is trained with a large learning rate.  This is consistent with a view that higher learning
rates lead to better high-level features, since, intuitively,
the features at the output layer are the most high-level.

{\bf Additional related previous work.}  The most closely related
previous work that we know of was described above.  Some recent papers
have highlighted limitations of NTK analyses
\cite{DBLP:conf/nips/WeiLLM19,DBLP:conf/nips/ChizatOB19,DBLP:conf/nips/YehudaiS19,DBLP:conf/nips/Allen-ZhuL19,DBLP:conf/colt/LiMZ20,DBLP:journals/corr/abs-2001-04413,DBLP:conf/nips/DanielyM20,DBLP:conf/nips/GhorbaniMMM20}.
A typical such
theoretical result is a learning problem that can be
solved (much) more efficiently using a neural network than by using
kernel method with the NTK.  Theoretical analyses with simple neural
network architectures in idealized settings motivate research on
natural data with practical architectures, identifying
qualitative differences between the features used before and after
training, demonstrating feature learning, and exploring its nature.  A
number of papers have described interesting properties of hidden units
in a neural network, including how they change during training (see
\cite{DBLP:journals/corr/Shwartz-ZivT17,DBLP:conf/nips/RaghuGYS17,saxe2019information,zhang2019all,DBLP:conf/nips/GolatkarAS19,DBLP:conf/iclr/FrankleC19,DBLP:conf/iclr/ZhangBHMS20,DBLP:conf/nips/NeyshaburSZ20,nguyen2021do,papyan2020prevalence}).
Because, in a precise sense, the after kernel captures the features
used by the network after training, its evolution during training is
especially interesting.  Lee, et al
\cite{DBLP:conf/nips/LeeSPAXNS20} compared learning with finite
networks to application of the NTK and the conjugate kernel, which may
be viewed as applying infinite networks.  They also evaluated the
effects of a number of modificiations of the training process.  Arora,
et al \cite{DBLP:conf/nips/AroraDH0SW19} provided an algorithm to
calculate the NTK for a convolutional neural network architecture, and
compared the application of this kernel with the training of
finite-width networks.  (See also
\cite{DBLP:conf/iclr/NovakXHLASS20}.)  Yang and Hu
\cite{yang2020feature} established conditions on the probability
distributions used to sample the initial weights, and the step size of
gradient descent, that imply that feature learning takes place, and
conditions that imply that it does not.  They also supported their
theoretical analysis with experiments.  Kornblith, et al
\cite{DBLP:conf/icml/Kornblith0LH19} proposed using centered kernel
alignment to measure the similarity of hidden nodes in the network; we
use kernel alignment here to compare the neural tangent kernel and its
conjugate projection.

{\bf Recent related work.}  After a preliminary version
of this paper was published on ArXiv \cite{long2021properties_v1},
\cite{ji2021fast} proposed new algorithms for finding the
maximum margin linear separator -- we use Algorithm 1 from
their paper in this version.  Recent papers evaluating
the evolution of the after kernel include \cite{shan2021rapid,atanasov2021neural}.

\section{Preliminaries and Methods}
\label{s:prelim}

We focus on two-class classification.


If $\theta$ is the vector of initial weights for a neural network
computing the function $f_{\theta}$, then 
{\em the finite
neural
tangent kernel} $K_{\mathrm{FNTK}}$ is defined by
$
K_{\mathrm{FNTK}}(u,v) = 
  (\nabla_{\theta} f_{\theta}(u)) \cdot (\nabla_{\theta} f_{\theta}(v)).
$
The {\em after kernel} is defined the same way after training.

A {\em layered neural network} takes as input $x$, and
transforms it using a processing layer $g_1$, to produce
a hidden layer $h_1(x) = g_1(x)$ of nodes, then applies another 
processing layer
$g_2$ to $h_1$, producing $h_2(x) = g_2(h_1(x))$, and so on.   
The final processing layer $g_L$, produces a scalar output 
$f(x) = g_L(h_{L-1}(x))$.

The {\em conjugate projection} $K_c$ is defined by
$K_c(u,v) = h_{L-1}(u) \cdot h_{L-1}(v)$.


Most of our experiments use networks with an architecture
similar to VGG \cite{SZ15}.  They are convolutional networks with
two blocks of layers, followed by a fully connected layer.
The first block consists of two convolutional layers with
26 channels each, $3 \times 3$ filters, and ReLU activations,
followed by a max-pooling layer with stride $2$ and pool size
$2$.  The second block is similar, except with $52$ channels.
We call these networks VGG-like networks.  (The number of
channels was chosen so that the number of parameters of the network
is approximately 100K, which is much more than the number of training
examples.  Additional experimental support for the proposition that
the networks in this paper are sufficiently
overparameterized may be found in Appendix~\ref{a:overparameterized}.
)

We also experiment with larger networks where the number
of channels in the convolutional layers are $119$ and
$238$ respectively.  We refer to these as the 
{\em mega-VGG-like networks} because they have approximately
one million parameters.

We also experiment with fully connected networks with four hidden
layers of width 94.  This was chosen so that the model has approximately
100K parameters.

Finally, for some unit tests, we experimented with a network
that simply computes the sum of its inputs, and passes the result through
fully connected layer with ReLU activations and width 16, followed by
a fully connected layer with one output.  
While this network is not useful for classification, and, as such, is not a object of 
study itself, it is translation and
rotation invariant, so it {\em is} useful for unit tests of the processes and 
code evaluating
these invariances; we used it for this.

These networks were implemented using Tensorflow 
\cite{tensorflow2015-whitepaper}
and Keras \cite{chollet2015}.  In
particular, for all the networks, all of the weights of all of the
layers were initialized using the default initializer for Keras.


In most of our experiments, training was conducted without
data augmentation.  When data augmentation was used,
it was implemented using the Keras 
ImageDataGenerator class, with a rotation range of 15
degrees, width shift range of 1, and height shift range of 1.


We trained the networks using the Keras implementation of
SGD, with the default batch size and step size.
We varied the number of epochs of training, using
values 0, 1, 3, 10, 30, 100, and 200.


We trained linear models on features extracted from the neural
networks.  
For this, we used Algorithm 1 from \cite{ji2021fast}.
Prior to linear training, the extracted features were rescaled so
that the average of the $L2$-norms of the features was one.  
(Note that, in principle, rescaling the features does not affect the 
output of the maximum margin algorithm.  In practice, however, the
rescaling was needed for numerical purposes.)


We experimented with three binary classification datasets:
(a) the 3 and 8 classes from MNIST;
(b) the 5 and 6 classes from MNIST, and
(c) the cat and dog classes from CIFAR10.


To evaluate the quality of kernels, we trained hard-margin
SVM classifiers 
on them 
and measured the test error.
We used all of the training data, both to train the deep networks
from which the kernels were extracted, and to train the SVM models
using those kernels.


To evaluate the translation invariance of a kernel
$K$ for each image $x$ in the test set, for
the embedding $\phi$ associated with $K$,
and for the images $s_h(x)$ and $s_v(x)$ obtained by
shifting $x$ one pixel to the right, and one pixel
down, we computed the cosine similarity with
$\phi(x)$ and $\phi(s_h(x))$, and the
cosine similarity with
$\phi(x)$ and $\phi(s_v(x))$.  We then average all of these 
similarities.


We measured rotation invariance similarly to translation invariance,
except rotating the images $1/4$ radians clockwise and
counterclockwise, instead of shifting them.


We measured invariance to zooms as follows.  
For each image $x$, we obtained a perturbed image
$z(x)$ cropping the outermost pixels from $x$, produced
a cropped image $c(x)$ with height and width two less than
$x$'s.  We then resized $c(x)$ to the original size of 
$x$, producing $z(x)$, using TensorFlow's resize command, with the
default parameters.  We computed the cosine similarity between
$x$ and $z(x)$, and average the result over all of the examples
in the test set.


We measured the extent to which a kernel is based on
local properties of an image as follows.  For each
image $x$ in the test set, we computed a perturbed image
$s(x)$ by swapping the upper left quadrant of $x$ with its
lower right quadrant.  We then computed the cosine similarity
between the embeddings of $x$ and $s(x)$ and averaged
over all examples in the test set.  
Note that the local statistics of $x$ and $s(x)$ are similar,
but, especially in MNIST, swapping the upper right and lower left
quadrants materially disrupts that global properties of the image.


We measured the kernel alignment \cite{NIPS2001_1f71e393,DBLP:conf/icml/Kornblith0LH19} 
between the NTK and its
conjugate projection, both before and after training.  The measure the
alignment between two kernels, we computed Gram matrices $G$ and $H$
for the respective kernels on the test data, and used $\frac{\vecrm(G)
  \cdot \vecrm(H)}{\sqrt{\vecrm(G) \cdot \vecrm(G)}\sqrt{\vecrm(H)
    \cdot \vecrm(H)}}$, where $\vecrm(G)$ and $\vecrm(H)$ are
flattenings of $G$ and $H$.

In addition to measuring the effect of the number of epochs of training of
the neural network on properties of the after kernel, we also measured the
effect of the learning rate used to train the neural network.  In these experiments, the neural network
was trained until the training loss was at most 0.01.  For the MNIST data sets,
we used learning rates 0.001, 0.003, 0.01, 0.03 and 0.1.  For the CIFAR10 data set,
we used learning rates 0.01, 0.02, 0.03, 0.04.  For each learning rate,
we evaluated the quality of the kernel, and its invariances, the same way as in the experiments 
that varied the number of epochs of neural network training. 


Every experiment was repeated 10 times for different random
initializations of the neural network weights, the results were
averaged, and confidence intervals were computed.

\section{Results}

\begin{figure}[tbp]
\hfill
\begin{subfigure}{\twocapwidth}
\centering
\includegraphics[width=\twofigwidth]{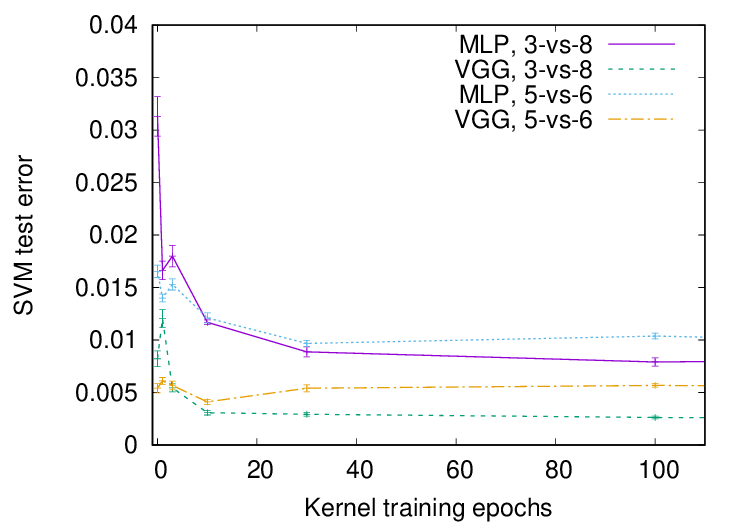}
\caption{The average test error of a hard margin SVM
on MNIST, as a function of the number of epochs of training
of the neural network used to produce the kernel.  Results
are plotted for fully connected networks, and VGG-like networks.}
\end{subfigure}
\hfill
\begin{subfigure}{\twocapwidth}
\centering
\includegraphics[width=\twofigwidth]{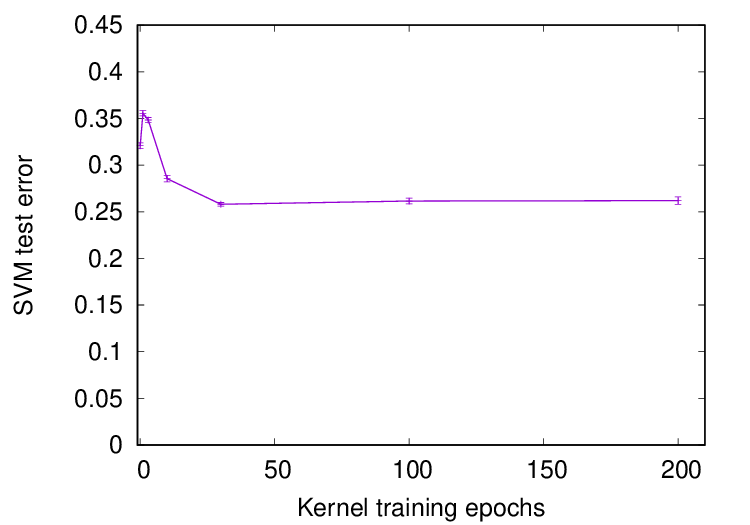}
\caption{The average test error of a hard margin SVM
on a CIFAR-10 cats-vs-dogs binary classification problem, 
as a function of the number of epochs of training
of the neural network used to produce the kernel, for the VGG-like network.}
\end{subfigure}
\hfill
\caption{}
\label{f:kernel_epochs_vs_svm_accuracy}
\end{figure}

Figure~\ref{f:kernel_epochs_vs_svm_accuracy}
plots the test error of the hard margin SVM using the after kernel,
as a function of the number of epochs that the network was trained
before it was used to determine the kernel.  
The FNTK corresponds to zero epochs of training.  When the kernel
is computed using a fully connected network, the kernel quickly
improves after a few epochs of training.  For the VGG networks on
the MNIST problems,
the FNTK was already pretty good, though some improvement
seems to be seen for the 3-vs-8 problem.  For the VGG network on
the CIFAR-10 problem, the after kernel is also much better than the
FNTK.

\begin{figure}[tbp]
\hfill
\begin{subfigure}{\twocapwidth}
\centering
\includegraphics[width=\twofigwidth]{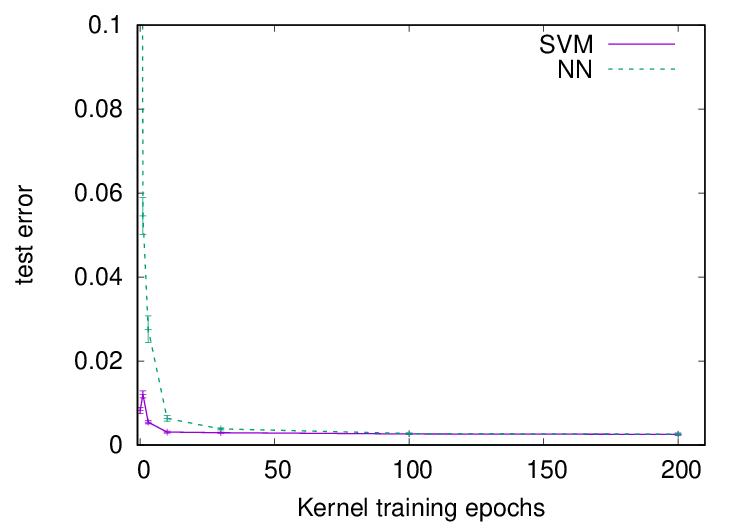}
\caption{
The plot for the VGG-like network
on MNIST 3-vs-8.
}
\label{f:kernel_epochs_vs_nn_svm_accuracy_vgg_mnist38_all}
\end{subfigure}
\hfill
\begin{subfigure}{\twocapwidth}
\centering
\includegraphics[width=\twofigwidth]{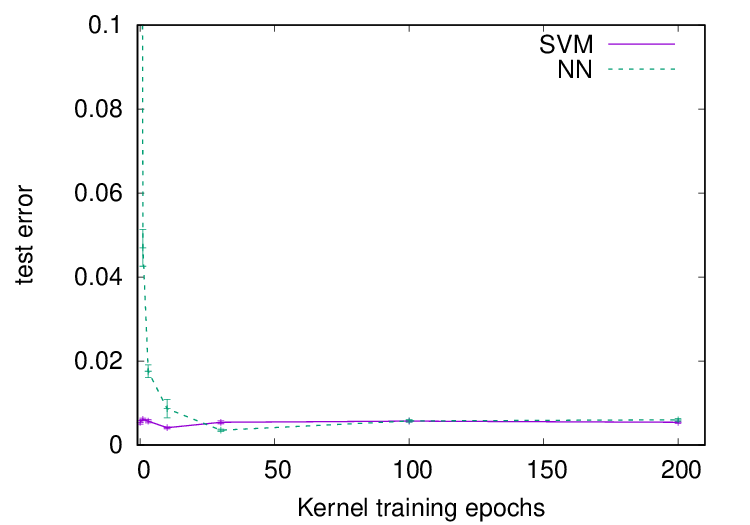}
\caption{
The plot for the VGG-like network
on MNIST 5-vs-6.
}
\label{f:kernel_epochs_vs_nn_svm_accuracy_vgg_mnist56_all}
\end{subfigure}
\hfill
\\
\hfill
\begin{subfigure}{\twocapwidth}
\centering
\includegraphics[width=\twofigwidth]{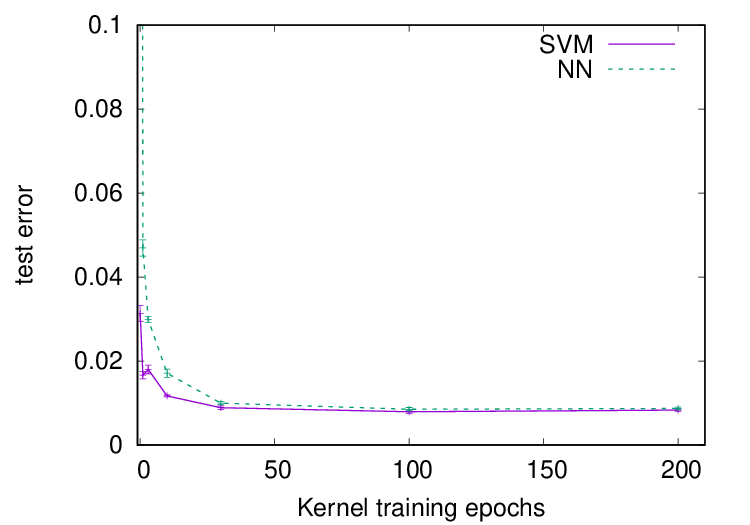}
\caption{
The plot for the fully connected network
on MNIST 3-vs-8.
}
\label{f:kernel_epochs_vs_nn_svm_accuracy_mlp_mnist38_all}
\end{subfigure}
\hfill
\begin{subfigure}{\twocapwidth}
\centering
\includegraphics[width=\twofigwidth]{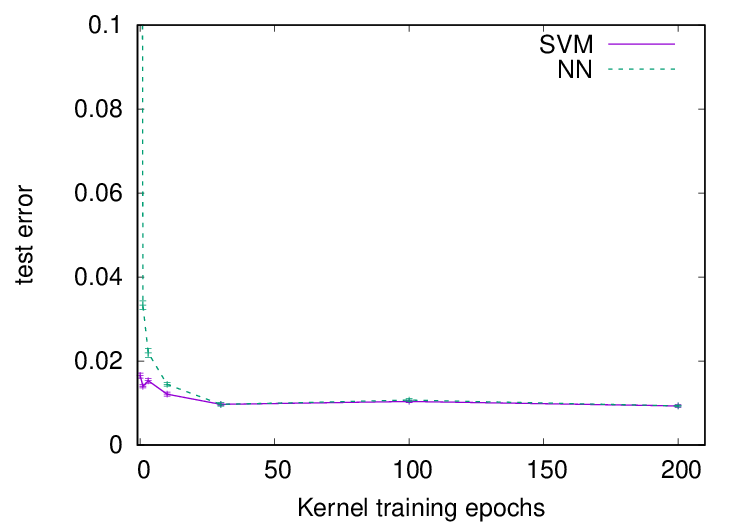}
\caption{
The plot for the fully connected network
on MNIST 5-vs-6.
}
\label{f:kernel_epochs_vs_nn_svm_accuracy_mlp_mnist56_all}
\end{subfigure}
\hfill
\\
\centering
\begin{subfigure}{\twocapwidth}
\centering
\includegraphics[width=\twofigwidth]{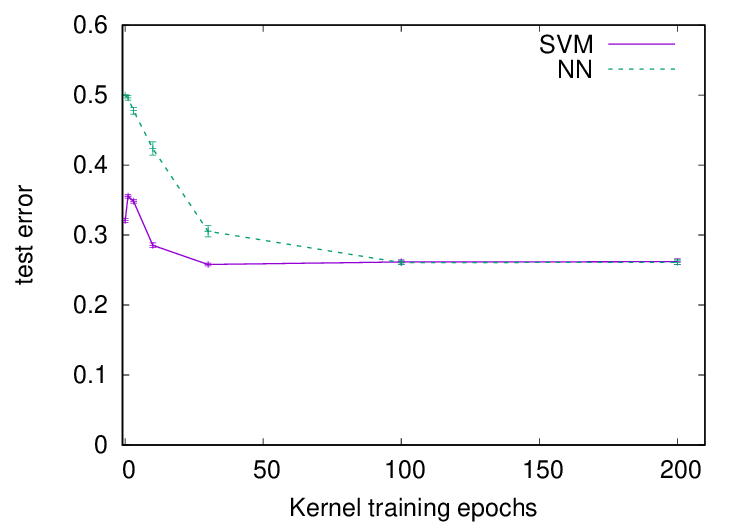}
\caption{The plot for the VGG-like network
on CIFAR10 cats-vs-dogs.
}
\label{f:kernel_epochs_vs_nn_svm_accuracy_vgg_cifar10_all}
\end{subfigure}
\hfill
\caption{Plots of the average test error of a hard margin SVM,
as a function of the number of epochs of training
of the neural network used to produce the kernel, along with the
test error of the neural network used to produce the kernel,
for different architecture/dataset pairs.}
\label{f:kernel_epochs_vs_nn_svm_accuracy}
\end{figure}
\begin{figure}[tbp]
\hfill
\begin{subfigure}{\twocapwidth}
\centering
\includegraphics[width=\twofigwidth]{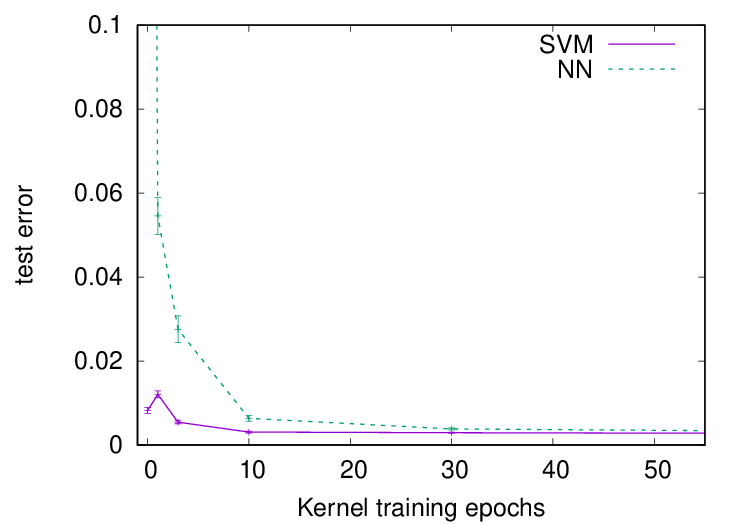}
\caption{
The plot for the VGG-like network
on MNIST 3-vs-8.
}
\label{f:kernel_epochs_vs_nn_svm_accuracy_vgg_mnist38_early}
\end{subfigure}
\hfill
\begin{subfigure}{\twocapwidth}
\centering
\includegraphics[width=\twofigwidth]{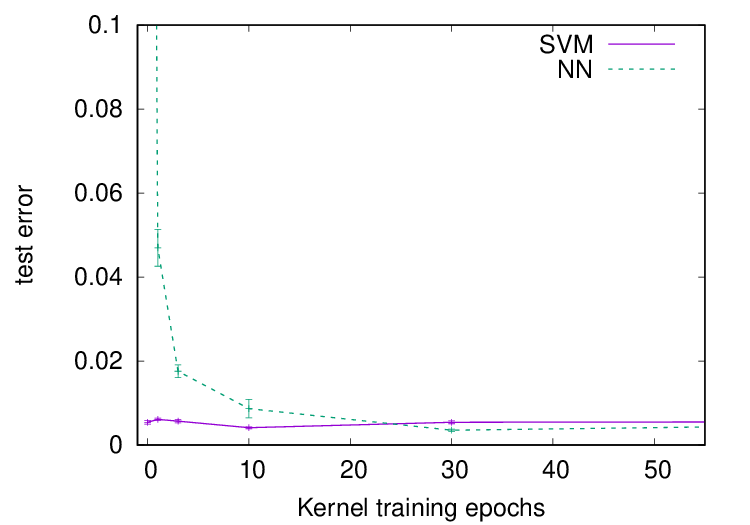}
\caption{
The plot for the VGG-like network
on MNIST 5-vs-6.
}
\label{f:kernel_epochs_vs_nn_svm_accuracy_vgg_mnist56_early}
\end{subfigure}
\hfill
\\
\hfill
\begin{subfigure}{\twocapwidth}
\centering
\includegraphics[width=\twofigwidth]{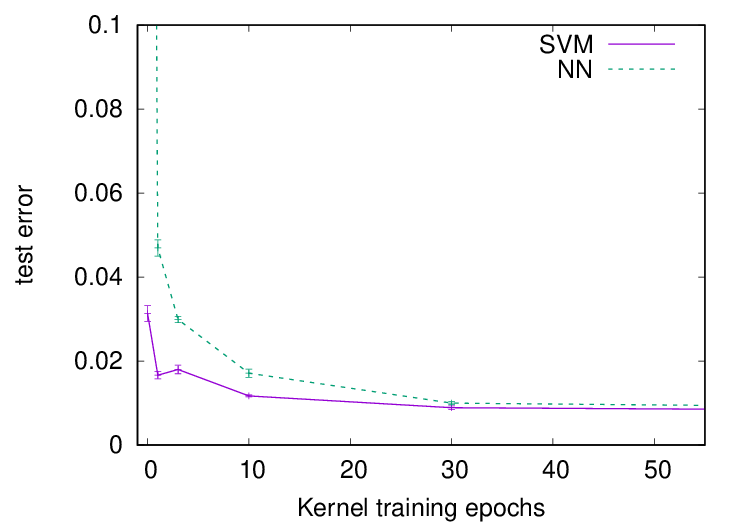}
\caption{
The plot for the fully connected network
on MNIST 3-vs-8.
}
\label{f:kernel_epochs_vs_nn_svm_accuracy_mlp_mnist38_early}
\end{subfigure}
\hfill
\begin{subfigure}{\twocapwidth}
\centering
\includegraphics[width=\twofigwidth]{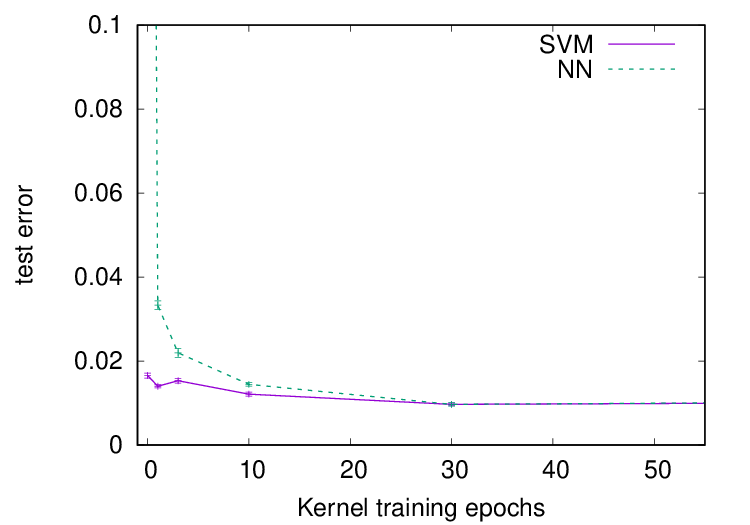}
\caption{
The plot for the fully connected network
on MNIST 5-vs-6.
}
\label{f:kernel_epochs_vs_nn_svm_accuracy_mlp_mnist56_early}
\end{subfigure}
\hfill
\\
\centering
\begin{subfigure}{\twocapwidth}
\centering
\includegraphics[width=\twofigwidth]{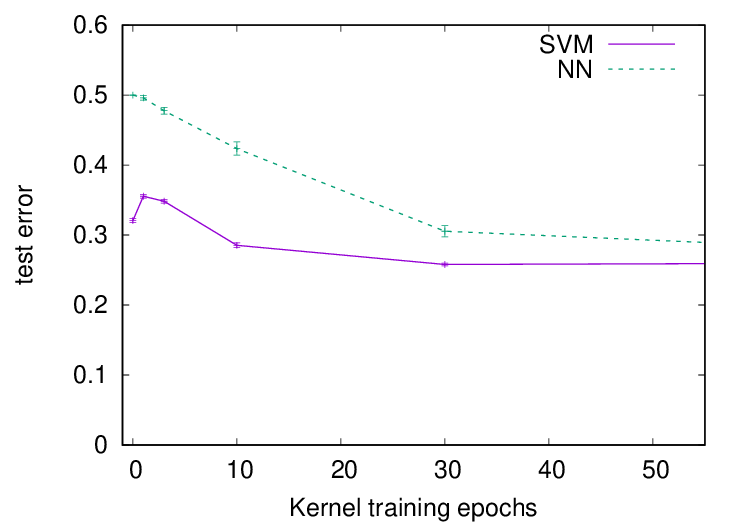}
\caption{The plot for the VGG-like network
on CIFAR10 cats-vs-dogs.
}
\label{f:kernel_epochs_vs_nn_svm_accuracy_vgg_cifar10_early}
\end{subfigure}
\hfill
\caption{Plots of the average test error of a hard margin SVM,
as a function of the number of epochs of training
of the neural network used to produce the kernel, along with the
test error of the neural network used to produce the kernel,
for different architecture/dataset pairs.}
\label{f:kernel_epochs_vs_nn_svm_accuracy_early}
\end{figure}

Figure~\ref{f:kernel_epochs_vs_nn_svm_accuracy}
once again plots to test error of the hard-margin SVM
with the after kernel, and also plots the test error of
the network used to produce the kernel.  
Figure~\ref{f:kernel_epochs_vs_nn_svm_accuracy_early}
provides the same plots, but concentrated on the portion of the curve
regarding the early stage of training.
The kernel is learned first, and then the network.
Ultimately, the test error of the neural network catches up
with the test error of the SVM trained with its after kernel,
consistent with the theory in \cite{LL20}.


\begin{figure}[tbp]
\hfill
\begin{subfigure}{\twocapwidth}
\centering
\includegraphics[width=\twofigwidth]{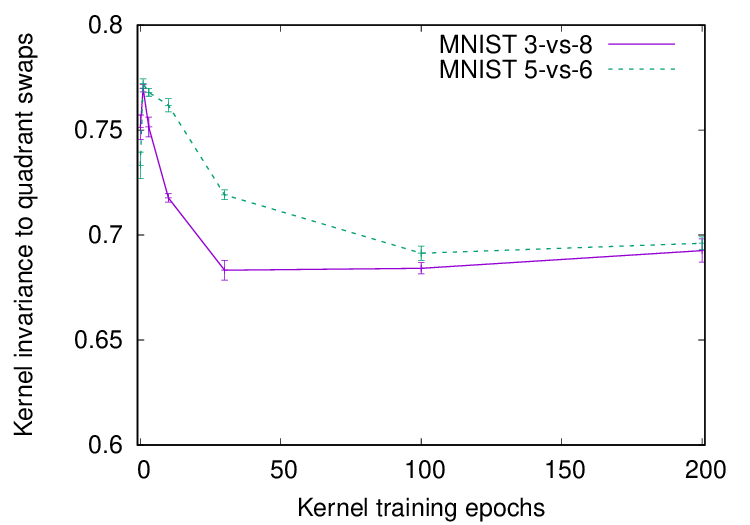}
\caption{The plot for VGG networks
on two MNIST two-class problems.}
\label{f:kernel_epochs_vs_swap_invariance_vgg}
\end{subfigure}
\hfill
\begin{subfigure}{\twocapwidth}
\centering
\includegraphics[width=\twofigwidth]{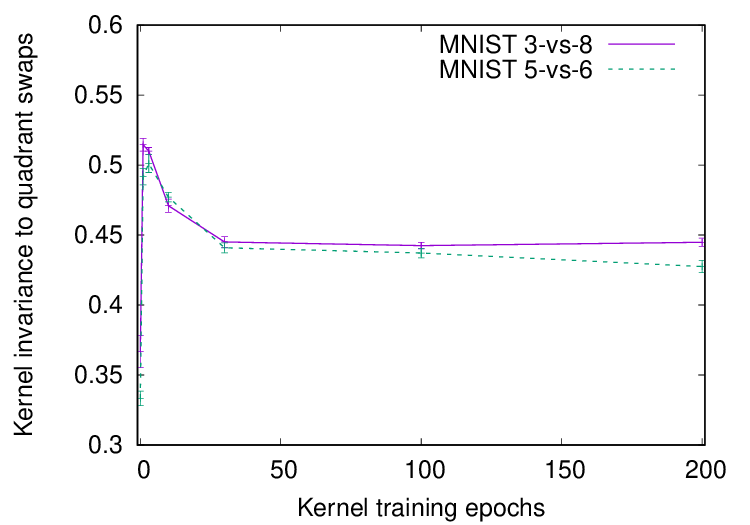}
\caption{The plot for fully
connected networks
on two MNIST two-class problems.}
\label{f:kernel_epochs_vs_swap_invariance_mlp}
\end{subfigure}
\hfill
\newline
\mbox{}
\hfill
\begin{subfigure}{\twocapwidth}
\centering
\includegraphics[width=\twofigwidth]{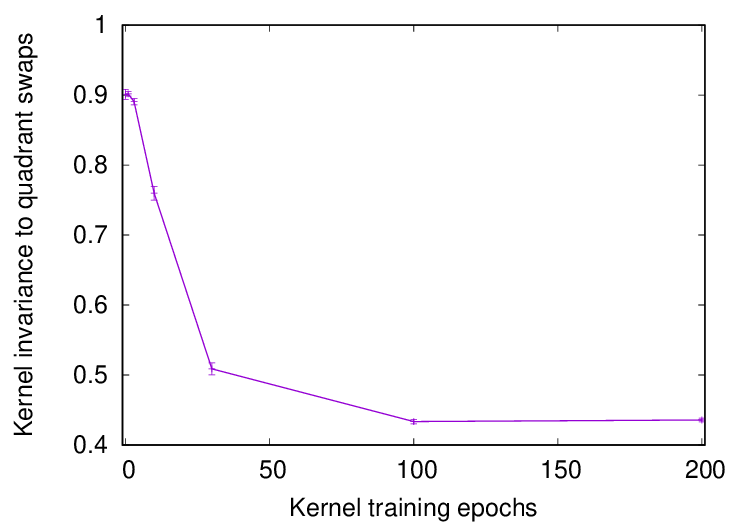}
\caption{The plot for the VGG-like
network on the CIFAR10 cats-vs-dogs problem.}
\label{f:kernel_epochs_vs_swap_invariance_cifar10}
\end{subfigure}
\hfill
\begin{subfigure}{\twocapwidth}
\centering
\includegraphics[width=\twofigwidth]{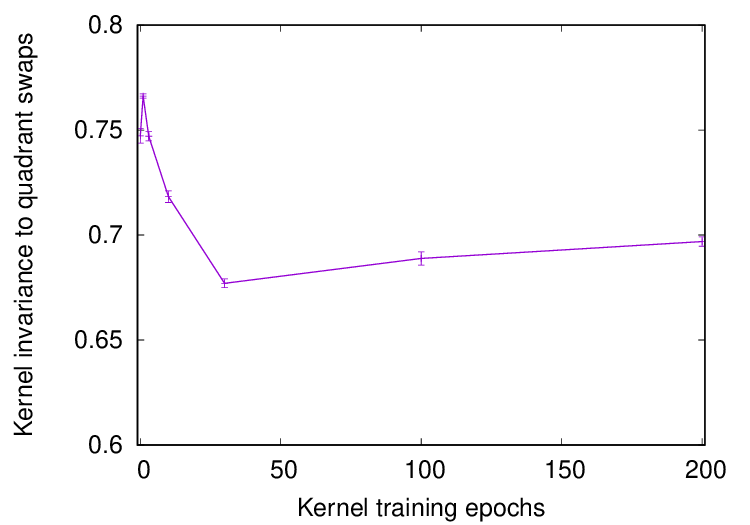}
\caption{The plot for the
mega-VGG-like network
on the MNIST 3-vs-8 problem.}
\label{f:kernel_epochs_vs_swap_invariance_mega}
\end{subfigure}
\hfill
\caption{Plots of the average cosine similarity between
features computed from the original image, and features
computed from an image with two quadrants swapped,
for the after kernel, as a function of the number
of epochs of training of the neural network used
to produce the kernel, for different architecture/dataset
pairs.  }
\label{f:kernel_epochs_vs_swap_invariance}
\end{figure}

Figure~\ref{f:kernel_epochs_vs_swap_invariance_vgg}
plots the invariance of the features associated with
the after kernel to swapping the top left quadrant
and the bottom right quadrant, for test images.
Recall that this is averaged over 10000 test examples
and 10 training runs.  These swaps largely preserve local
statistics, so the reduction seen in
Figure~\ref{f:kernel_epochs_vs_swap_invariance_vgg}
in the insensitivity to these swaps is an indication that
the kernel depends more on global properties of the image.

Figure~\ref{f:kernel_epochs_vs_swap_invariance_mlp} is the analogous
plot for fully connected networks on the MNIST 3-vs-8 problem.  The
decreasing trend in swap-invariance with the number of rounds of
training is consistent with the results for the VGG networks.
However, the swap-invariance at initialization is less, and is the
least ever seen.  This could be because, for fully connected networks
at initialization, many of the features are nonsense, informally,
computing a function that is a locality sensitive hash of the input
image.  A non-local perturbation of the image then produces unrelated
features.  Overall, the curves are consistent with the possibility
that, in fully connected networks, local features are learned first,
then global features.

Figure~\ref{f:kernel_epochs_vs_swap_invariance_cifar10}
is the analogous plot for the VGG-like network on
the CIFAR10 cats-vs-dogs problem.  The swap-invariance
decreases substantially, showing that the kernel depends
more on global properties of the image.  The effect may be
greatest with the CIFAR10 because global information is 
needed to a greater extent to differentiate between the classes.

Figure~\ref{f:kernel_epochs_vs_swap_invariance_mega}
is the analogous plot for the mega-VGG-like network.  
For this bigger network, the after-kernel
has significantly less swap invariance.

\begin{figure}[tbp]
\hfill
\begin{subfigure}{\twocapwidth}
\centering
\includegraphics[width=\twofigwidth]{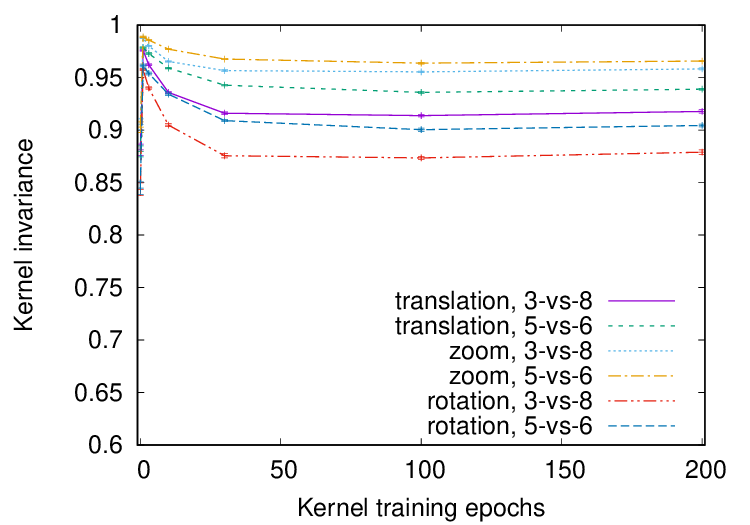}
\caption{The plot for VGG networks
on two MNIST two-class problems.}
\label{f:kernel_epochs_vs_tzr_invariances_vgg}
\end{subfigure}
\hfill
\begin{subfigure}{\twocapwidth}
\centering
\includegraphics[width=\twofigwidth]{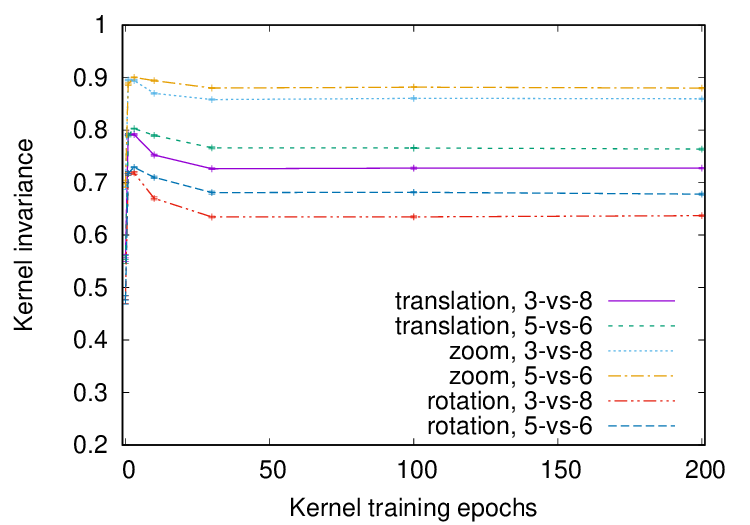}
\caption{The plot for fully
connected networks
on two MNIST two-class problems.}
\label{f:kernel_epochs_vs_tzr_invariances_mlp}
\end{subfigure}
\hfill
\newline
\mbox{}
\hfill
\begin{subfigure}{\twocapwidth}
\centering
\includegraphics[width=\twofigwidth]{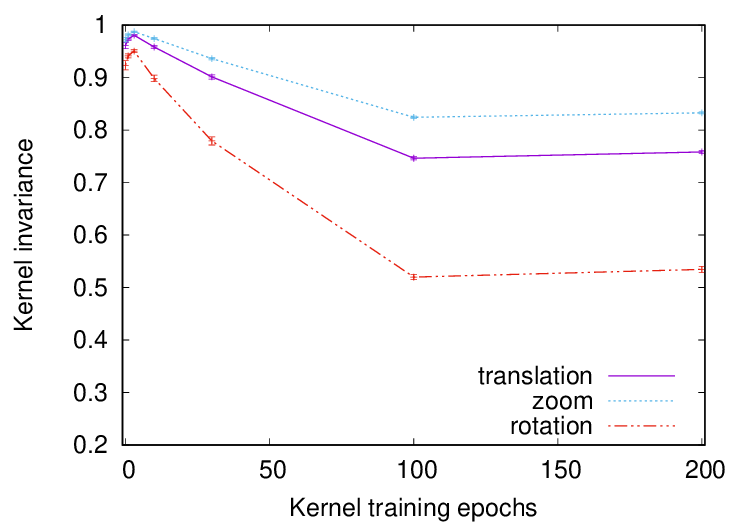}
\caption{The plot for the VGG-like
network on the CIFAR10 cats-vs-dogs problem.}
\label{f:kernel_epochs_vs_tzr_invariances_cifar10}
\end{subfigure}
\hfill
\begin{subfigure}{\twocapwidth}
\centering
\includegraphics[width=\twofigwidth]{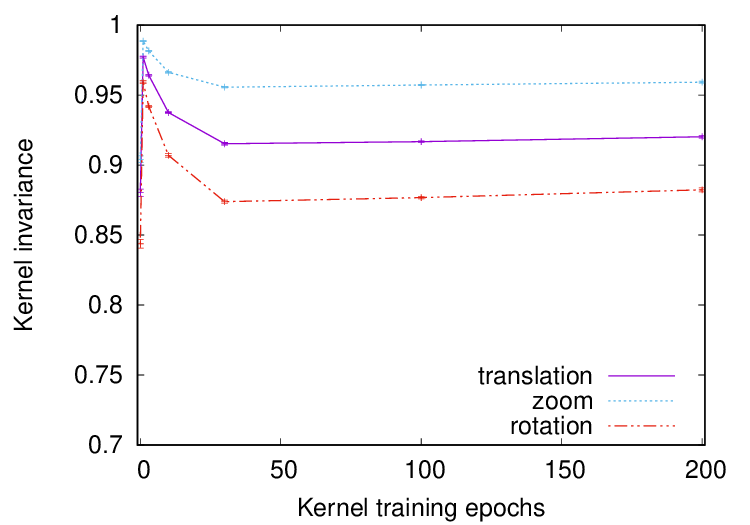}
\caption{The plot for the
mega-VGG-like network
on the MNIST 3-vs-8 problem.}
\label{f:kernel_epochs_vs_tzr_invariances_mega}
\end{subfigure}
\hfill
\caption{Plots of the average cosine similarity between
features computed from the original image, and features
computed from an image with various perturbations,
for the after kernel, as a function of the number
of epochs of training.  The perturbations include
shifts by one pixel, slight zooms, and rotations by
small angles.  The networks include the VGG-like network,
the mega-VGG-like network, and the fully connected network.  The
datasets include MNIST 3-vs-8 and 5-vs-6 problems, and the
CIFAR10 cats-vs-dogs problem.}
\label{f:kernel_epochs_vs_tzr_invariances}
\end{figure}

Figure~\ref{f:kernel_epochs_vs_tzr_invariances}
plots how invariant kernels are to different perturbations
of the inputs, for different combinations of architectures,
data sets, and number of rounds of training.

In all of the cases in the plot, the invariance shoots up after one
epoch of training, then decreases as training continues.  In most cases, the
invariance settles down to a value larger than the FNTK.  The fully
connected network's kernel starts out less invariant, and continues to be
throughout training.  The after kernel for the VGG-like network on
CIFAR10 is much less rotation invariant than at initialization.


\begin{figure}[tbp]
\hfill
\begin{subfigure}{\threecapwidth}
\centering
\includegraphics[width=\threefigwidth]{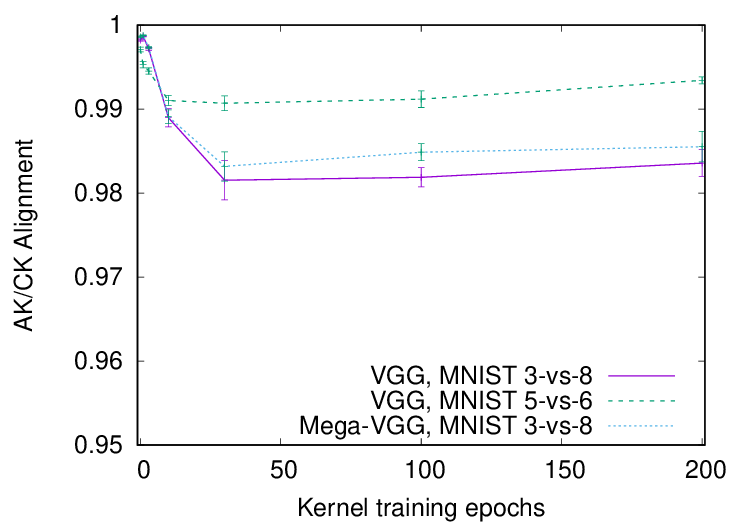}
\caption{Alignments between
the after kernel and its conjugate projection for
VGG-like networks.}
\label{f:kernel_epochs_vs_ak_ck_alignment_vgg}
\end{subfigure}
\hfill
\begin{subfigure}{\threecapwidth}
\centering
\includegraphics[width=\threefigwidth]{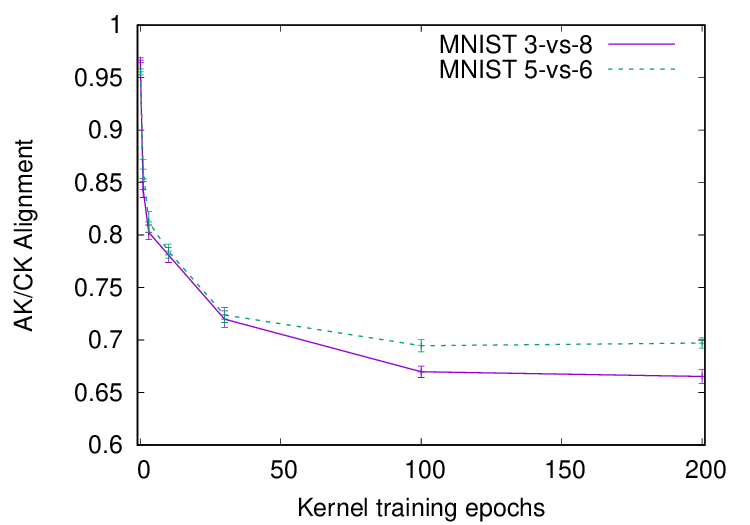}
\caption{Alignment between
the after kernel and its conjugate projection for
fully connected networks on MNIST problems.}
\label{f:kernel_epochs_vs_ak_ck_alignment_mlp}
\end{subfigure}
\hfill
\begin{subfigure}{\threecapwidth}
\centering\includegraphics[width=\threefigwidth]{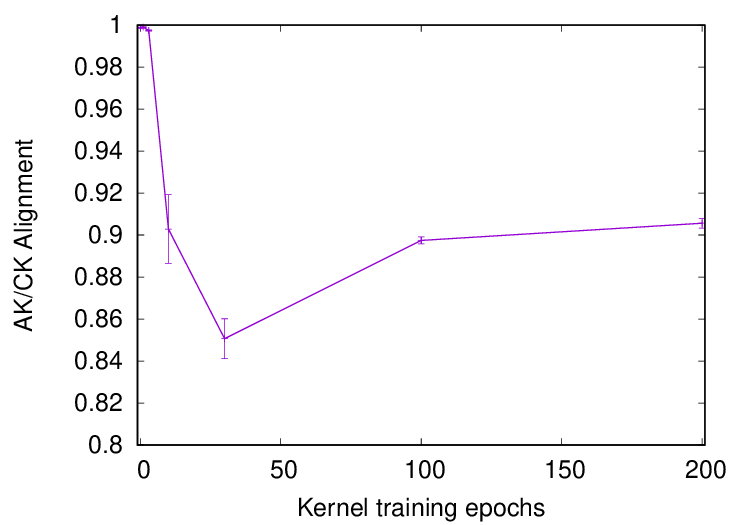}
\caption{Alignment between
after kernel and its conjugate projection for the VGG-like network on
cats-vs-dogs.}
\label{f:kernel_epochs_vs_ak_ck_alignment_cifar10}
\end{subfigure}
\hfill
\caption{Alignments between the after kernel and the conjugate kernel.}
\label{f:ak_ck_alignment}
\end{figure}

\begin{figure}[tbp]
\hfill
\begin{subfigure}{\twocapwidth}
\centering
\includegraphics[width=\twofigwidth]{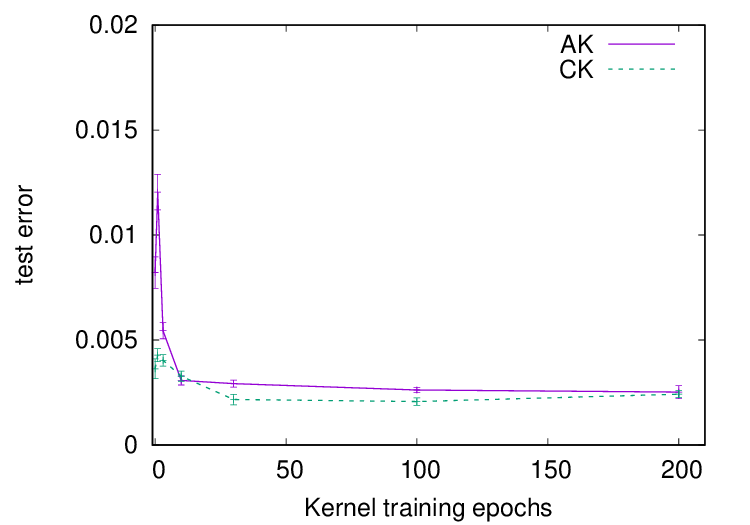}
\caption{The plot for
VGG-like networks on MNIST 3-vs-8.}
\label{f:kernel_epochs_vs_ak_ck_accuracy_vgg_mnist38}
\end{subfigure}
\hfill
\begin{subfigure}{\twocapwidth}
\centering
\includegraphics[width=\twofigwidth]{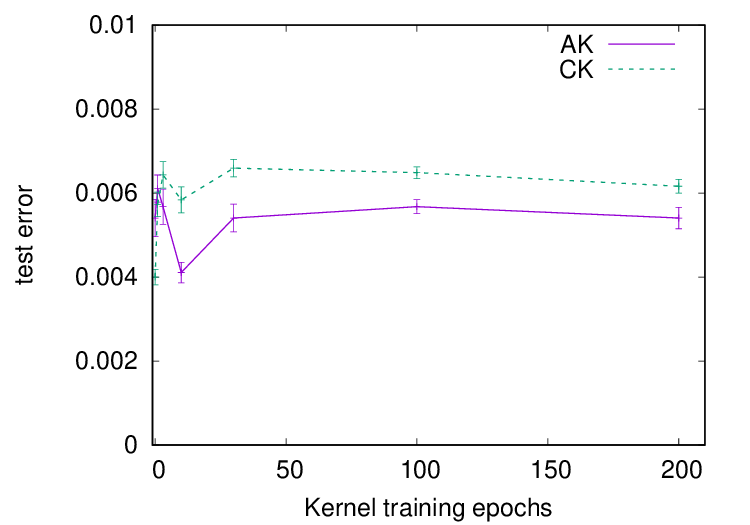}
\caption{The plot for
VGG-like networks on MNIST 5-vs-6.}
\label{f:kernel_epochs_vs_ak_ck_accuracy_vgg_mnist56}
\end{subfigure}
\hfill \\
\hfill
\begin{subfigure}{\twocapwidth}
\centering
\includegraphics[width=\twofigwidth]{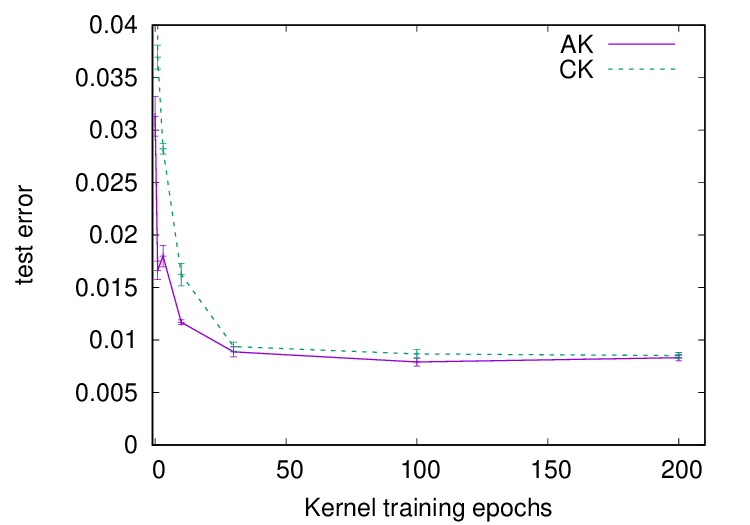}
\caption{The plot for
fully connected networks on MNIST 3-vs-8.}
\label{f:kernel_epochs_vs_ak_ck_accuracy_mlp_mnist38}
\end{subfigure}
\hfill
\begin{subfigure}{\twocapwidth}
\centering
\includegraphics[width=\twofigwidth]{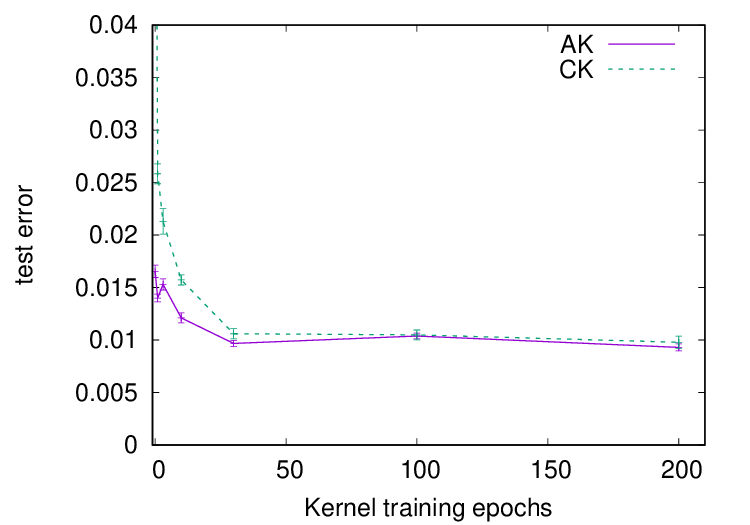}
\caption{The plot for
fully connected networks on MNIST 5-vs-6.}
\label{f:kernel_epochs_vs_ak_ck_accuracy_mlp_mnist56}
\end{subfigure}
\hfill \\
\centering
\hfill
\begin{subfigure}{\twocapwidth}
\centering
\includegraphics[width=\twofigwidth]{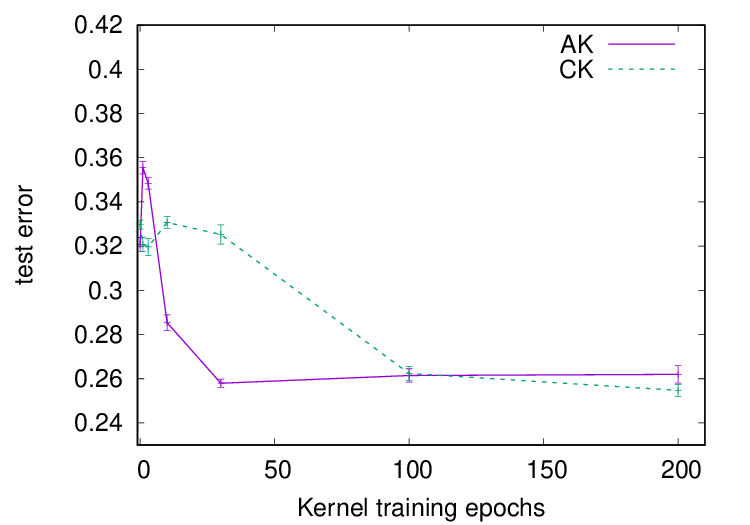}
\caption{The plot for the VGG-like network on
cats-vs-dogs.}
\label{f:kernel_epochs_vs_ak_ck_accuracy_vgg_cifar10}
\end{subfigure}
\hfill \\
\caption{Test error of SVMs trained with the after kernel and the conjugate kernel
for various architecture-dataset pairs.}
\label{f:ak_ck_accuracy}
\end{figure}

Figure~\ref{f:kernel_epochs_vs_ak_ck_alignment_vgg}
plots
the alignments between the after kernel and its conjugate projection,
for the VGG-like and mega-VGG-like networks, after different training
times.  The conjugate projection, which can be computed much more
efficiently, remains a good approximation throughout, but is an
especially good approximation at initialization.
Figure~\ref{f:kernel_epochs_vs_ak_ck_alignment_mlp}
plots
the alignments between the after kernel and its conjugate projection, for the fully connected networks, as a function
of the number of epochs of training.  The conjugate projection is
a good approximation at initialization, but becomes much less so
after training.
Figure~\ref{f:kernel_epochs_vs_ak_ck_alignment_cifar10}
plots
the alignments between the after kernel and 
its conjugate projection, for the VGG-like network, as a function
of the number of epochs of training.  Here again, the conjugate projection is
a good approximation at initialization, but becomes much less so
after training.

Figure~\ref{f:ak_ck_accuracy}
plots the test error of hard-margin SVM classifiers using
the after kernel and its conjugate projection, as a function
of the number of epochs used to train the kernels.  In most cases,
the classifier computed using the conjugate kernel is nearly
as accurate as the classifier that uses the after kernel.  It appears
that the features closest to the output layer are the most useful.
An exception, however, is the case of VGG networks on MNIST 5-vs-6,
where the hard-margin classifier using the after kernel is substantially
more accurate.

\begin{figure}[tbp]
\hfill
\begin{subfigure}{\twocapwidth}
\centering
\includegraphics[width=\twofigwidth]{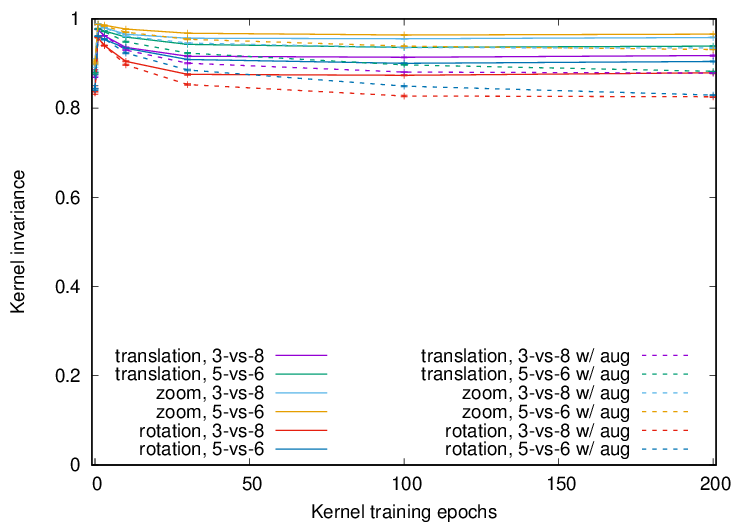}
\caption{The plot for VGG networks
on two MNIST two-class problems.}
\label{f:kernel_epochs_vs_tzr_invariances_vgg_w_aug}
\end{subfigure}
\hfill
\begin{subfigure}{\twocapwidth}
\centering
\includegraphics[width=\twofigwidth]{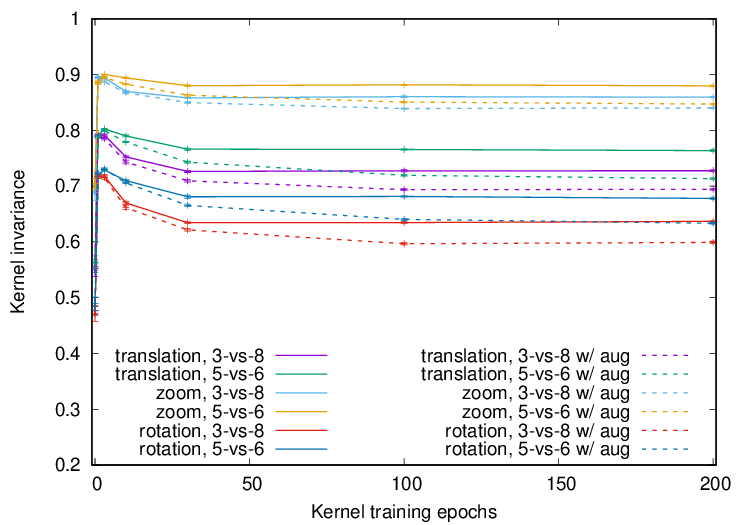}
\caption{The plot for fully
connected networks
on two MNIST two-class problems.}
\label{f:kernel_epochs_vs_tzr_invariances_mlp_w_aug}
\end{subfigure}
\hfill
\newline
\mbox{}
\hfill
\begin{subfigure}{\twocapwidth}
\centering
\includegraphics[width=\twofigwidth]{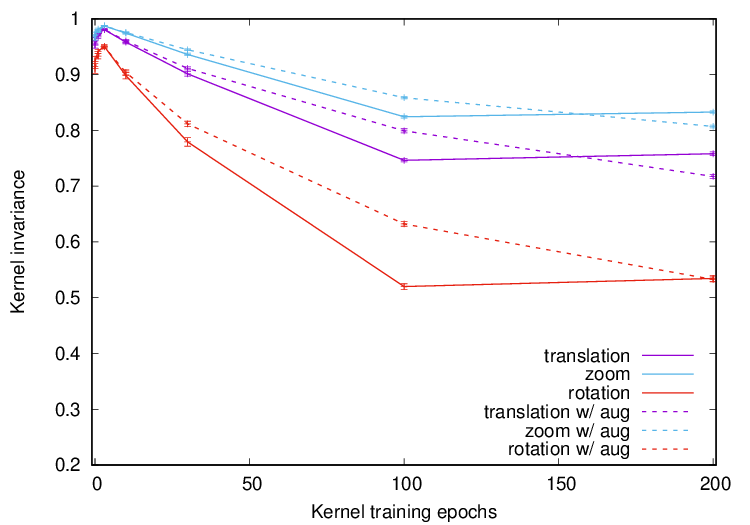}
\caption{The plot for the VGG-like
network on the CIFAR10 cats-vs-dogs problem.}
\label{f:kernel_epochs_vs_tzr_invariances_cifar10_w_aug}
\end{subfigure}
\hfill
\begin{subfigure}{\twocapwidth}
\centering
\includegraphics[width=\twofigwidth]{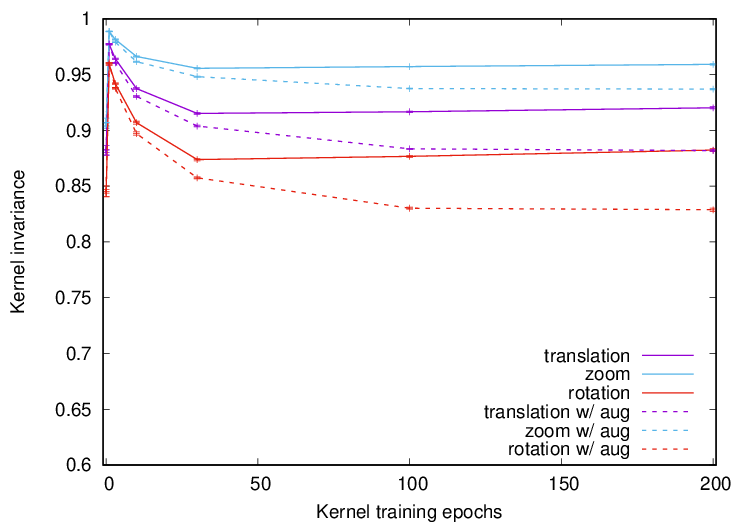}
\caption{The plot for the
mega-VGG-like network
on the MNIST 3-vs-8 problem.}
\label{f:kernel_epochs_vs_tzr_invariances_mega_w_aug}
\end{subfigure}
\hfill
\caption{Plots of three invariances of kernels with and without data
  augmentation, as a function of the number of epochs of training.}
\label{f:kernel_epochs_vs_tzr_invariances_w_aug}
\end{figure}

Figure~\ref{f:kernel_epochs_vs_tzr_invariances_w_aug} 
plots the invariances of the kernels to
translations, zooms and rotations, when models are trained both with
and without data augmentation.

As a sanity check, 
Figure~\ref{f:kernel_epochs_vs_aug_noaug_accuracies}
in Appendix~\ref{a:sanity}
reproduces the well-known fact that data augmentation
improves the test accuracy of trained models, and that this
carries over to the models trained with SVM using kernels
trained with data augmentation.

\begin{figure}[tbp]
\hfill
\begin{subfigure}{\twocapwidth}
\centering
\includegraphics[width=\twofigwidth]{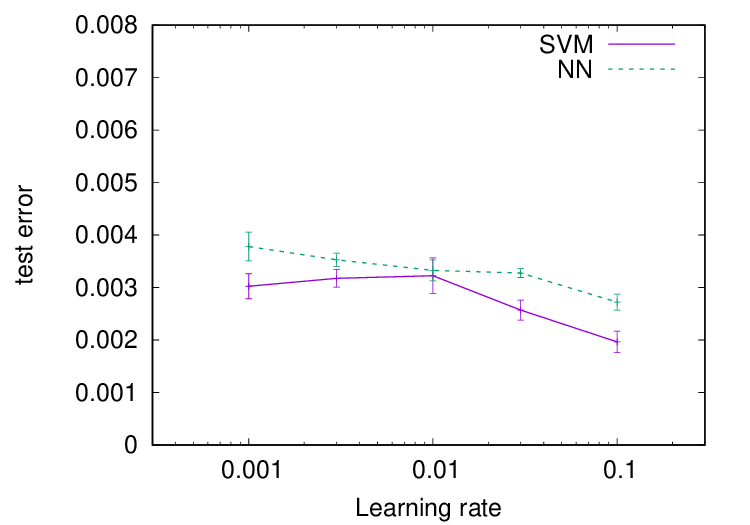}
\caption{
The plot for the VGG-like network
on MNIST 3-vs-8.
}
\label{f:lr_vs_nn_svm_accuracy_vgg_mnist38}
\end{subfigure}
\hfill
\begin{subfigure}{\twocapwidth}
\centering
\includegraphics[width=\twofigwidth]{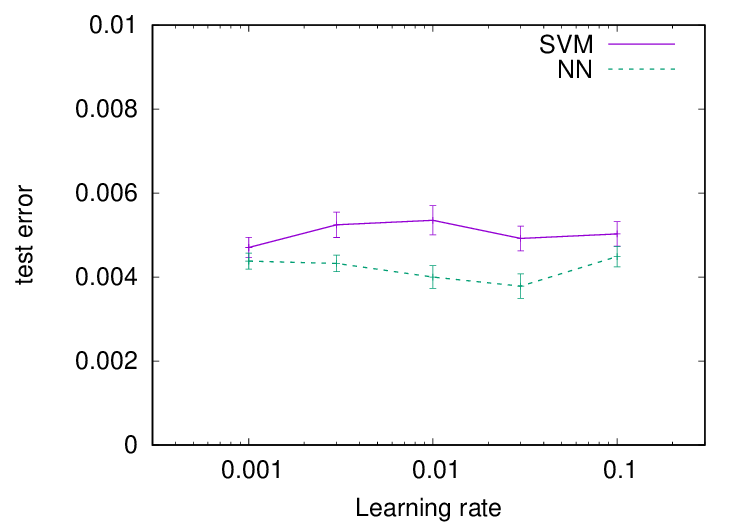}
\caption{
The plot for the VGG-like network
on MNIST 5-vs-6.
}
\label{f:lr_vs_nn_svm_accuracy_vgg_mnist56}
\end{subfigure}
\hfill
\\
\hfill
\begin{subfigure}{\twocapwidth}
\centering
\includegraphics[width=\twofigwidth]{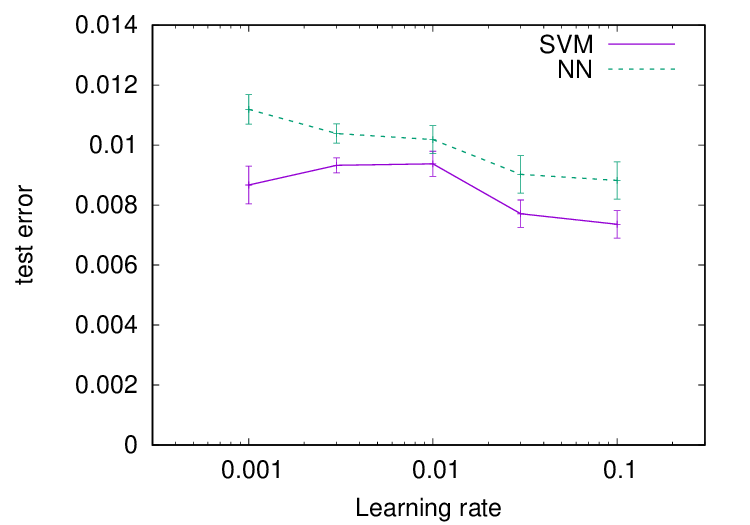}
\caption{
The plot for the fully connected network
on MNIST 3-vs-8.
}
\label{f:lr_vs_nn_svm_accuracy_mlp_mnist38}
\end{subfigure}
\hfill
\begin{subfigure}{\twocapwidth}
\centering
\includegraphics[width=\twofigwidth]{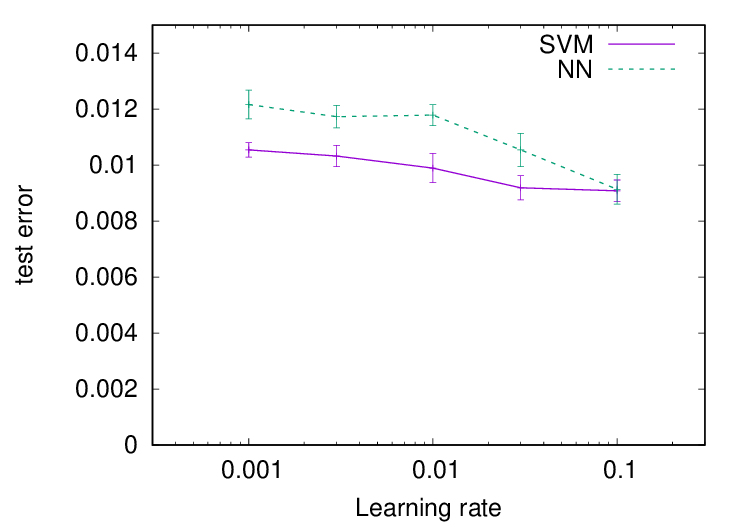}
\caption{
The plot for the fully connected network
on MNIST 5-vs-6.
}
\label{f:lr_vs_nn_svm_accuracy_mlp_mnist56}
\end{subfigure}
\hfill
\\
\centering
\begin{subfigure}{\twocapwidth}
\centering
\includegraphics[width=\twofigwidth]{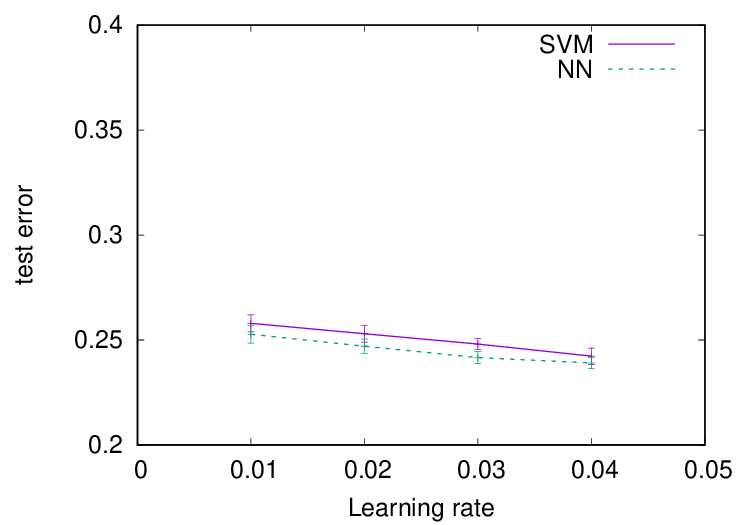}
\caption{The plot for the VGG-like network
on CIFAR10 cats-vs-dogs.
}
\label{f:lr_vs_nn_svm_accuracy_vgg_cifar10}
\end{subfigure}
\hfill
\caption{Plots of the average test error of a hard margin SVM,
as a function of the learning rate when
the neural network used to produce the kernel, along with the
test error of the neural network used to produce the kernel,
for different architecture/dataset pairs.}
\label{f:lr_vs_nn_svm_accuracy}
\end{figure}

Figure~\ref{f:lr_vs_nn_svm_accuracy} plots the accuracy
of the maximum margin classifier using the after kernel,
along with the accuracy of the neural network used to
produce the kernel, as a function of the learning rate.
There is a general trend that a higher learning rate produces
a better kernel, if the training error is held constant.
On MNIST 3-vs-8,
the maximum margin classifier had better test error than the neural
network, despite the fact that the neural network was trained to almost
zero error.  Clearly, here, the neural network is not approximating the
maximum margin with respect to the after kernel.

\begin{figure}[tbp]
\hfill
\begin{subfigure}{\twocapwidth}
\centering
\includegraphics[width=\twofigwidth]{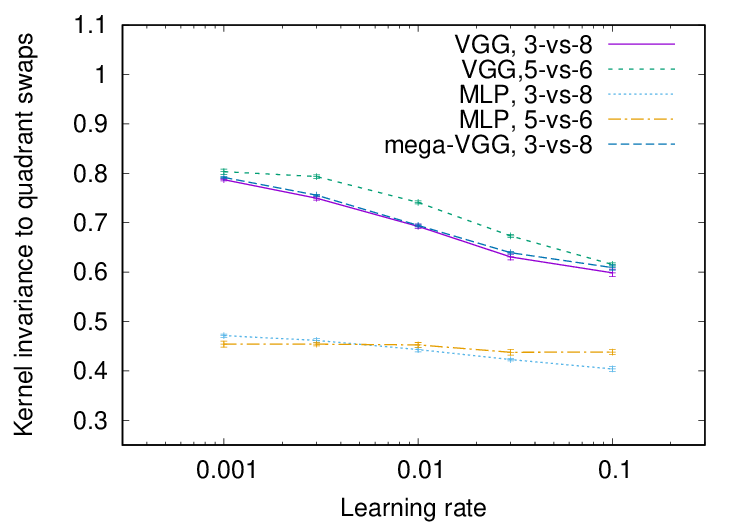}
\caption{The plot for  MNIST two-class problems.}
\label{f:lr_vs_swap_invariance_mnist}
\end{subfigure}
\hfill
\begin{subfigure}{\twocapwidth}
\centering
\includegraphics[width=\twofigwidth]{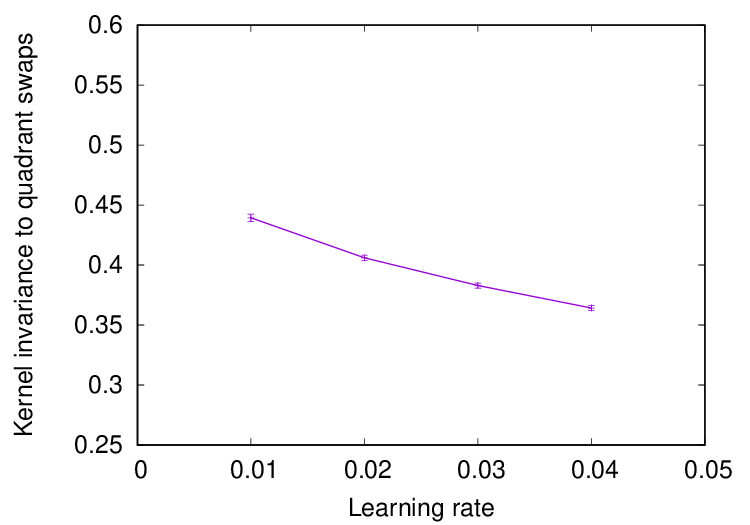}
\caption{The plot for VGG networks on CIFAR10 cats-vs-dogs.}
\label{f:lr_vs_swap_invariance_mlp}
\end{subfigure}
\hfill
\caption{Plots of the average cosine similarity between
features computed from the original image, and features
computed from an image with two quadrants swapped,
for the after kernel, as a function of learning rate, for different architecture/dataset
pairs.  }
\label{f:lr_vs_swap_invariance}
\end{figure}

Figure~\ref{f:lr_vs_swap_invariance} plots the swap invariance as a function
of the learning rate.  The features obtained with a large learning rate
reflect global properties of the image to a greater extent.

\begin{figure}[tbp]
\hfill
\begin{subfigure}{\twocapwidth}
\centering
\includegraphics[width=\twofigwidth]{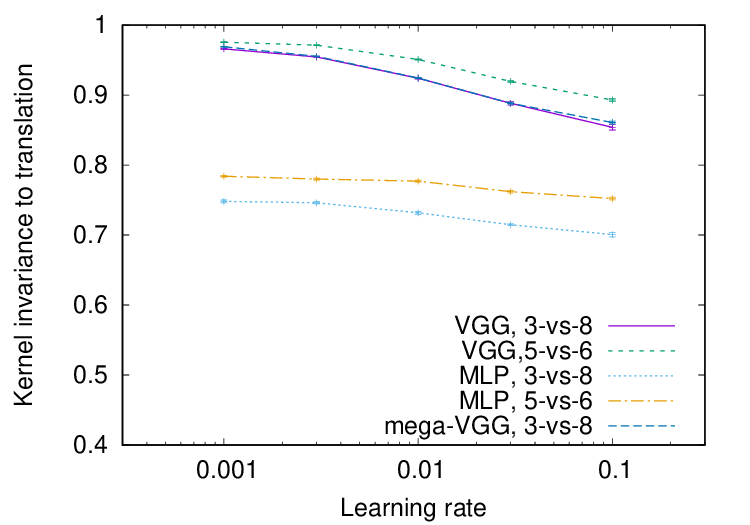}
\caption{The plot for translation invariance for MNIST two-class problems.}
\label{f:lr_vs_translation_invariance_mnist}
\end{subfigure}
\hfill
\begin{subfigure}{\twocapwidth}
\centering
\includegraphics[width=\twofigwidth]{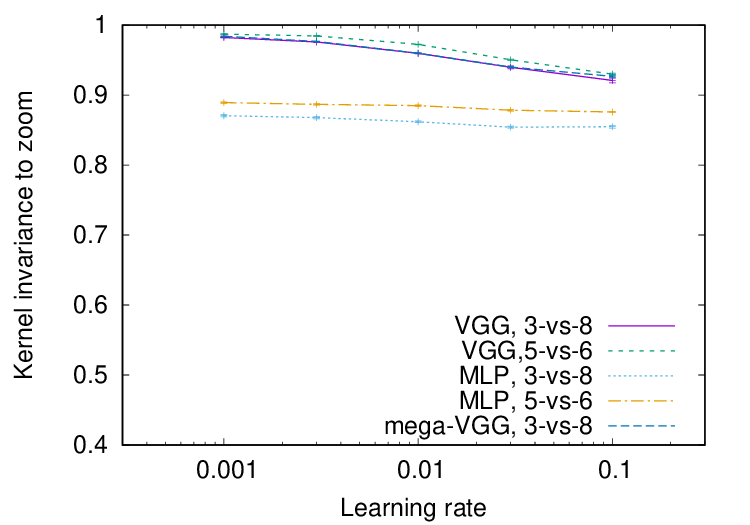}
\caption{The plot for zoom invariance for MNIST two-class problems.}
\label{f:lr_vs_zoom_invariance_mnist}
\end{subfigure}
\hfill
\newline
\hfill
\begin{subfigure}{\twocapwidth}
\centering
\includegraphics[width=\twofigwidth]{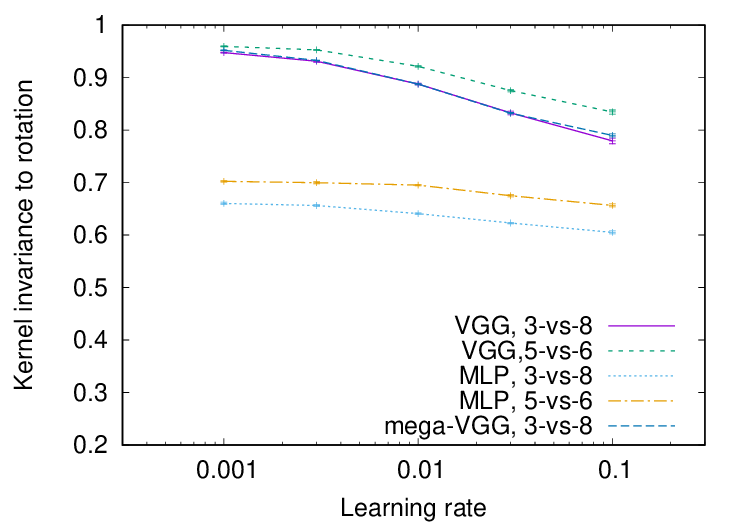}
\caption{The plot for rotation invariance for MNIST two-class problems.}
\label{f:lr_vs_rotation_invariance_mnist}
\end{subfigure}
\hfill
\begin{subfigure}{\twocapwidth}
\centering
\includegraphics[width=\twofigwidth]{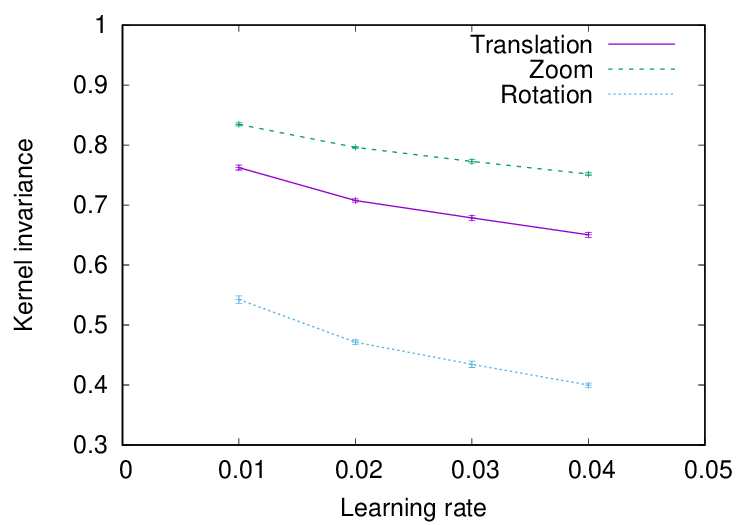}
\caption{The plot for VGG networks on CIFAR10 cats-vs-dogs.}
\label{f:lr_vs_tzr_invariance_cifar10}
\end{subfigure}
\hfill
\caption{Plots of translation, zoom, and rotation invariance, 
as a function of learning rate, for different architecture/dataset
pairs.  }
\label{f:lr_vs_tzr_invariance}
\end{figure}

Figure~\ref{f:lr_vs_tzr_invariance} plots the translation, room and rotation invariances as a function
of the learning rate.  Features obtained for larger learning rates are
less invariant to these slight transformations of the image.

\begin{figure}[tbp]
\hfill
\begin{subfigure}{\twocapwidth}
\centering
\includegraphics[width=\twofigwidth]{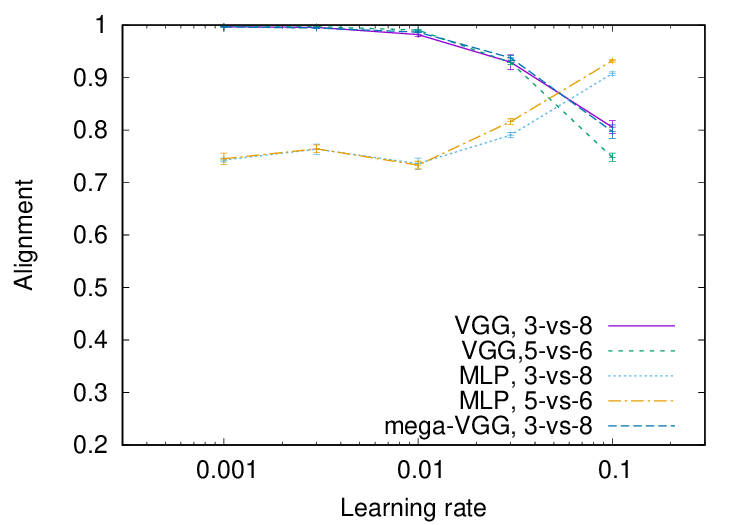}
\caption{The plot for  MNIST problems.}
\label{f:lr_vs_ak_ck_alignment_vgg}
\end{subfigure}
\hfill
\begin{subfigure}{\twocapwidth}
\centering\includegraphics[width=\twofigwidth]{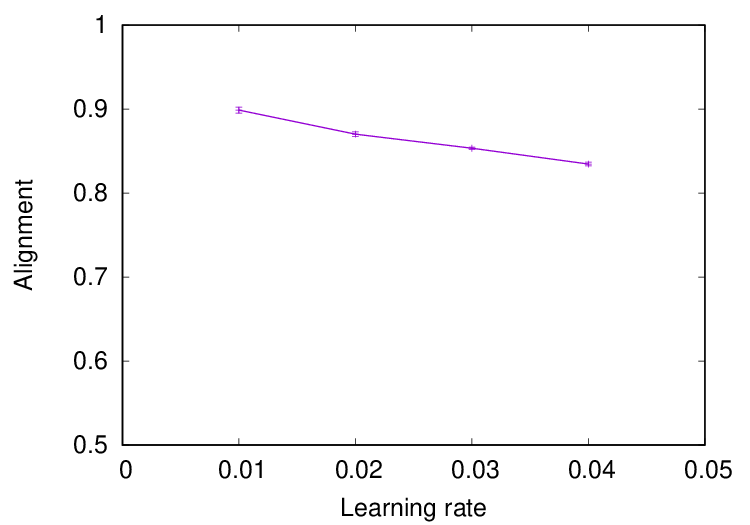}
\caption{The plot for CIFAR10
cats-vs-dogs.}
\label{f:lr_vs_ak_ck_alignment_cifar10}
\end{subfigure}
\hfill
\caption{Alignments between the after kernel and the conjugate kernel.}
\label{f:lr_vs_ak_ck_alignment}
\end{figure}

Figure~\ref{f:lr_vs_ak_ck_alignment} plots the alignment between
the after kernel and the conjugate kernel, as a function of
the learning rate.  For VGG networks, the alignment is highest
at the lowest learning rates, while the opposite holds for 
fully connected networks.

\begin{figure}[tbp]
\hfill
\begin{subfigure}{\twocapwidth}
\centering
\includegraphics[width=\twofigwidth]{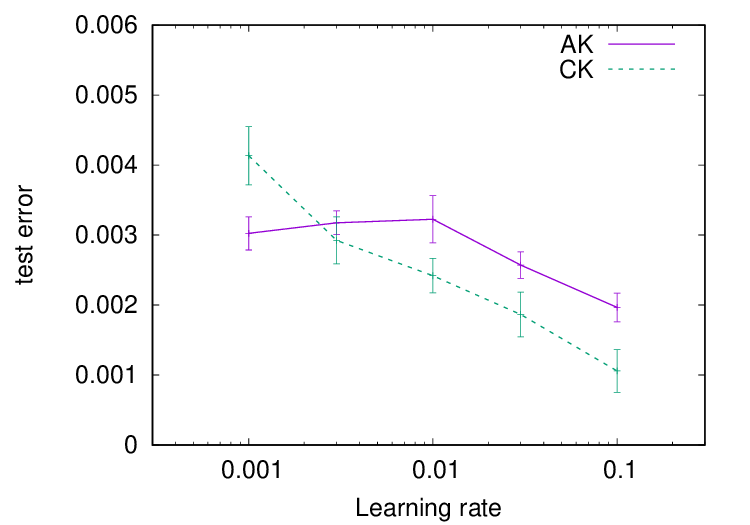}
\caption{The plot for
VGG-like networks on MNIST 3-vs-8.}
\label{f:lr_vs_ak_ck_accuracy_vgg_mnist38}
\end{subfigure}
\hfill
\begin{subfigure}{\twocapwidth}
\centering
\includegraphics[width=\twofigwidth]{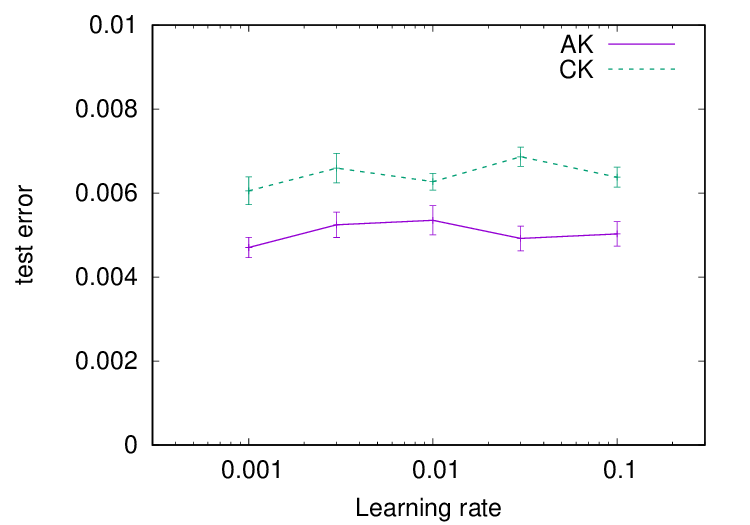}
\caption{The plot for
VGG-like networks on MNIST 5-vs-6.}
\label{f:lr_vs_ak_ck_accuracy_vgg_mnist56}
\end{subfigure}
\hfill \\
\hfill
\begin{subfigure}{\twocapwidth}
\centering
\includegraphics[width=\twofigwidth]{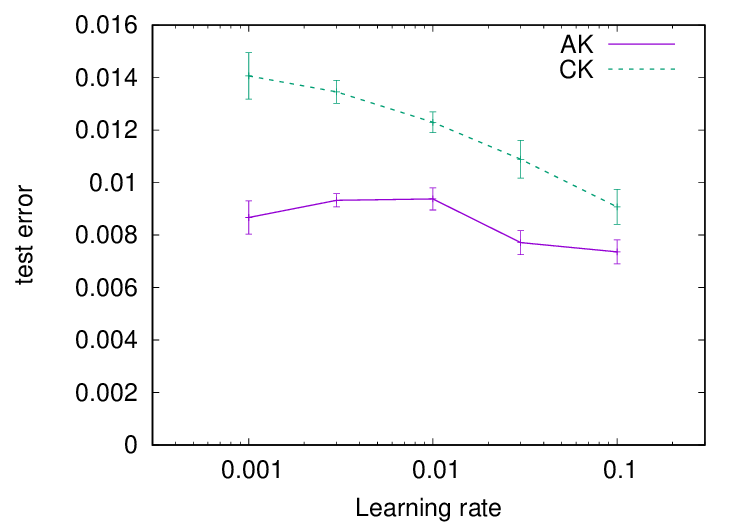}
\caption{The plot for
fully connected networks on MNIST 3-vs-8.}
\label{f:lr_vs_ak_ck_accuracy_mlp_mnist38}
\end{subfigure}
\hfill
\begin{subfigure}{\twocapwidth}
\centering
\includegraphics[width=\twofigwidth]{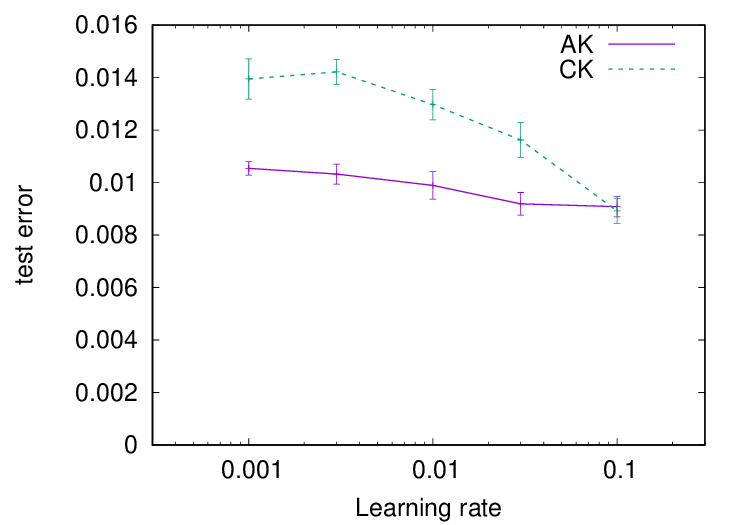}
\caption{The plot for
fully connected networks on MNIST 5-vs-6.}
\label{f:lr_vs_ak_ck_accuracy_mlp_mnist56}
\end{subfigure}
\hfill \\
\centering
\hfill
\begin{subfigure}{\twocapwidth}
\centering
\includegraphics[width=\twofigwidth]{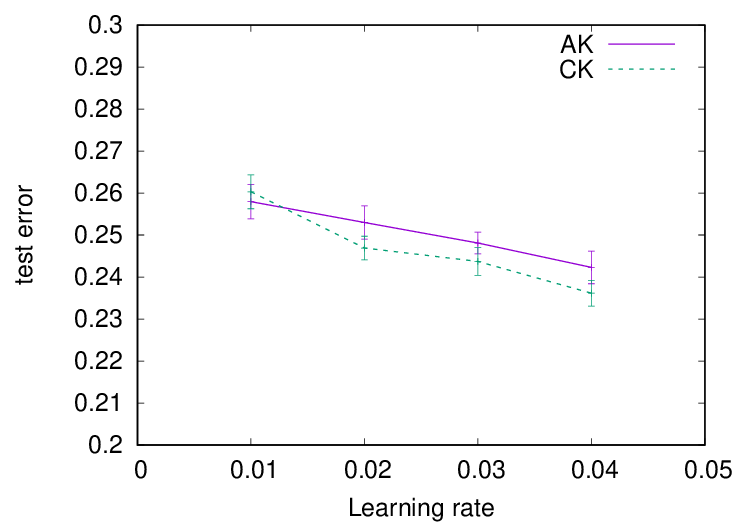}
\caption{The plot VGG-like networks on CIFAR10
cats-vs-dogs.}
\label{f:lr_vs_ak_ck_accuracy_vgg_cifar10}
\end{subfigure}
\hfill \\
\caption{Test error of SVMs trained with the after kernel and the conjugate kernel,
as a function of learning rate.}
\label{f:lr_vs_ak_ck_accuracy}
\end{figure}

Figure~\ref{f:lr_vs_ak_ck_accuracy} plots the test error of
maximum margin classifiers using the after and the conjugate kernel.
Generally, the after kernel provides better accuracy, though
the conjugate kernel is more competitive overall at higher
learning rates -- this is consistent with a view that higher learning
rates lead to better high-level features, since, intuitively,
the features at the output layer are the most high-level.

\begin{figure}[tbp]
\hfill
\begin{subfigure}{\twocapwidth}
\centering
\includegraphics[width=\twofigwidth]{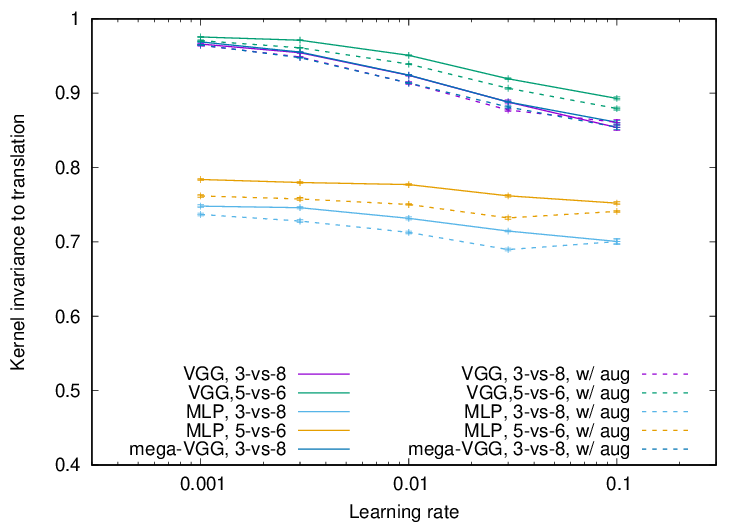}
\caption{Translation invariance for MNIST two-class problems.}
\label{f:lr_vs_translation_invariance_mnist_w_aug}
\end{subfigure}
\hfill
\begin{subfigure}{\twocapwidth}
\centering
\includegraphics[width=\twofigwidth]{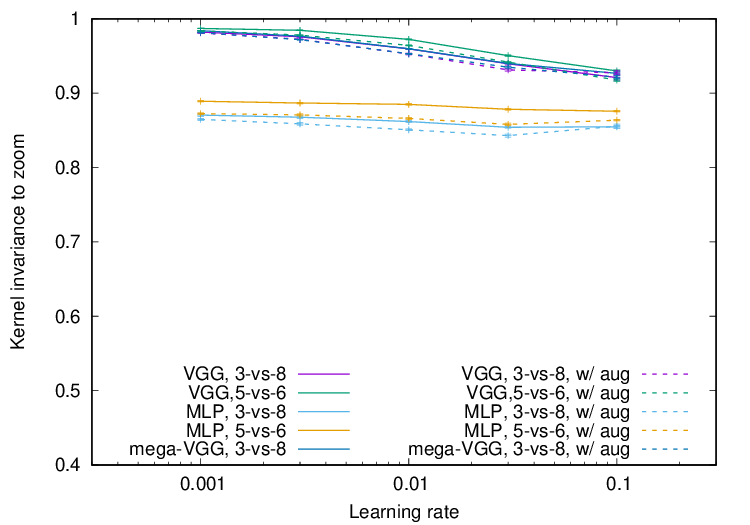}
\caption{Zoom invariance for MNIST two-class problems.}
\label{f:lr_vs_zoom_invariance_mnist_w_aug}
\end{subfigure}
\hfill
\newline
\hfill
\begin{subfigure}{\twocapwidth}
\centering
\includegraphics[width=\twofigwidth]{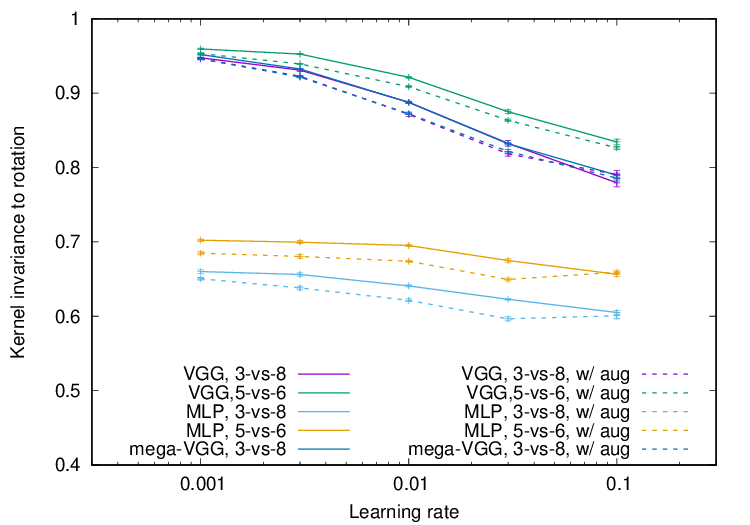}
\caption{Rotation invariance for MNIST two-class problems.}
\label{f:lr_vs_rotation_invariance_mnist_w_aug}
\end{subfigure}
\hfill
\begin{subfigure}{\twocapwidth}
\centering
\includegraphics[width=\twofigwidth]{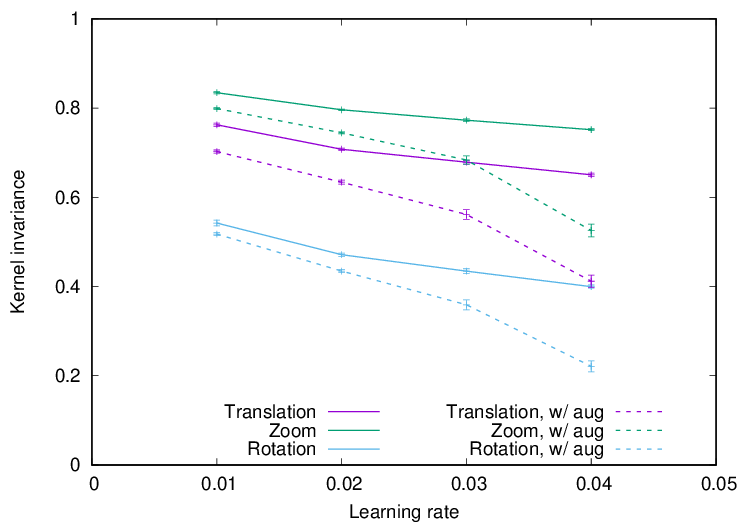}
\caption{The plot for VGG networks on CIFAR10 cats-vs-dogs.}
\label{f:lr_vs_tzr_invariance_cifar10_w_aug}
\end{subfigure}
\hfill
\caption{Plots of translation, zoom, and rotation invariance, 
as a function of learning rate, for different architecture/dataset
pairs, for data with and without augmentations.  }
\label{f:lr_vs_tzr_invariance_w_aug}
\end{figure}

Figure~\ref{f:lr_vs_tzr_invariance_w_aug} plots the translation, room
and rotation invariances as a function of the learning rate, for data
with augmentations.  The trend that a higher learning rate leads to less
invariance to these perturbations persists when augmentations are added,
and is even greater in the case of CIFAR10.

\section{Discussion}

When standard neural networks are trained in a standard way,
the after kernel is often qualitatively quite different from
the (finite) neural tangent kernel.  The quality of the
after kernel improves rapidly; it appears that the kernel is
learned first, then the kernel classifier.  Training a
network with a larger learning rate produces a better kernel,
even when the training error is held constant.  The advantage
of the kernel obtained with a larger learning rate can be
seen both in the accuracy of an SVM trained with the kernel,
and in invariance properties that meaningful features should
be expected to have.

The kernel used with the network after training is not strongly
invariant to small perturbations of the inputs.  We were surprised
that, in many cases, the degree of invariance degrades slowly
in the later stages of training.  Explaining this is a challenge
for theory.  Data augmentation did not make the after kernel
more invariant to small perturbations.

Many additional questions could be addressed by modifying the code
used in this research, which, as mentioned earlier, is available
online \cite{Lon21akcode}.  For example, it would be interesting to
evaluate the effect of contrastive training \cite{chen2020simple} on
the evolution of the after kernel, in particular, on its
invariances.  Comparing properties of the after
kernel for networks trained with EfficientNet
\cite{tan2019efficientnet} with networks trained with VGG could
enhance understanding of why EfficientNet models are more accurate.
Similarly, recent encouraging results with Visual Transformers
\cite{parmar2018image,bello2019attention,DBLP:conf/iclr/DosovitskiyB0WZ21}
motivate study of the after kernel of these models.  Examining the
effect of the depth of the network is another obvious future
direction.  Comparing the dynamics when models are trained for NLP
tasks with the dynamics arising from image data would be another
interesting subject for study.  Last, but not least, results on the effect of
the scale of the initialization on the implicit bias of gradient
descent in idealized theoretical settings
\cite{geiger2020disentangling,DBLP:conf/colt/WoodworthGLMSGS20}
motivate investigation of this effect on practical networks with real
data; the code accompanying this paper \cite{Lon21akcode} could be a
useful tool for this.




\section*{Acknowledgements}

We are grateful to
Yamini Bansal,
Peter Bartlett, 
Olivier Bousquet,
Yuan Cao,
Niladri Chatterji,
Quanquan Gu,
Ziwei Ji,
Praneeth Netrapalli,
Behnam Neyshabur,
Hanie Sedghi,
and
Jascha Sohl-Dickstein
valuable conversations.
We also thank Ziwei Ji for code for Algorithm 1 from \cite{ji2021fast}.

\bibliographystyle{plain}
\bibliography{bib}


\appendix

\section{The networks are sufficiently overparameterized}
\label{a:overparameterized}

In this appendix, we describe experiments aimed as supporting
the proposition that the networks in our study of the
after kernel are sufficiently
overparameterized.  We varied the width of the networks, while
otherwise holding the architecture fixed, as described in
Section~\ref{s:prelim}.
In the case of the 
VGG-like networks, this was done by scaling down the number 
of channels in each convolutional layer, and, in the fully connected
networks, this was done by reducing the size of the hidden layers.
For both architectures, we evaluated smaller networks with
approximately 60K and 80K parameters, along with the networks
used in our main studies, which had roughly 100K parameters.
For the three datasets used in our study, we evaluated the
test error of the neural networks of various sizes, trained using
the protocol described in Section~\ref{s:prelim},
including averaging over ten runs.

\begin{figure}[tbh]
\begin{subfigure}{\textwidth}
\centering
\includegraphics[width=4in]{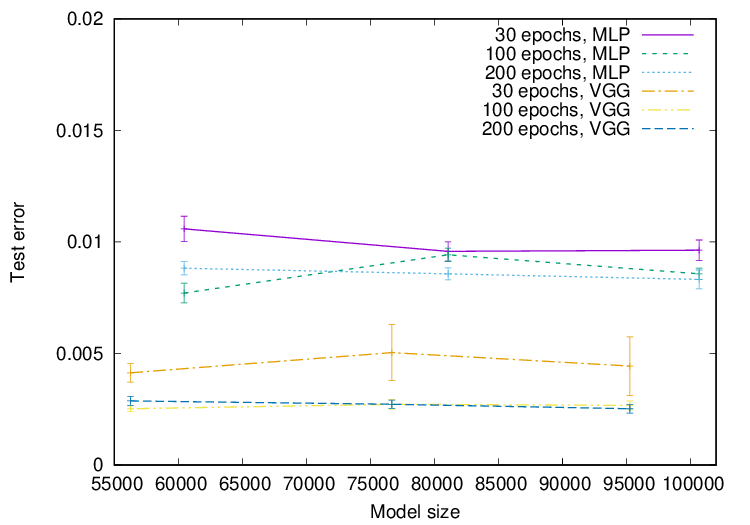}
\caption{Plots for MNIST 3-vs-8 and 5-vs-6.}
\end{subfigure}
\newline
\begin{subfigure}{\textwidth}
\centering
\includegraphics[width=4in]{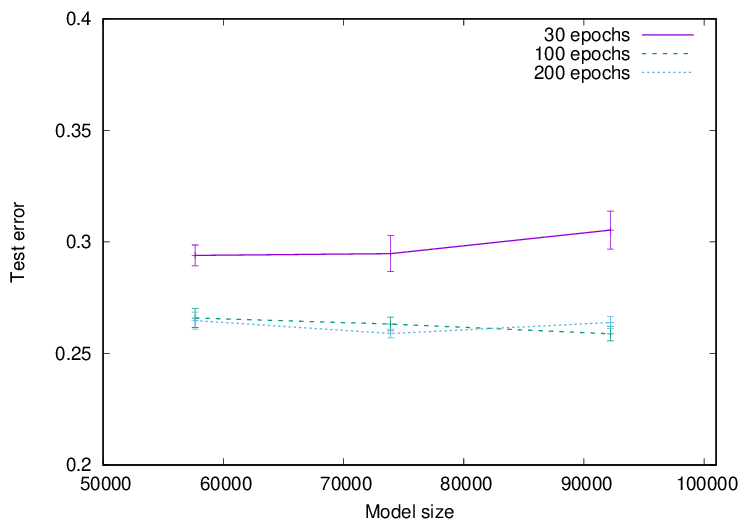}
\caption{Plots for CIFAR10 cats-vs-dogs}
\end{subfigure}
\caption{A plot of the average test error of neural networks
as a function of the parameters in the network, for
different architectures and numbers of epochs of training.}
\label{f:size_and_epochs_vs_accuracy}
\end{figure}

Figure~\ref{f:size_and_epochs_vs_accuracy}
plots the average test errors, as a function of the size of the
model.  In all cases, the test error does not improve much as
the size of the model increases from $\approx 60K$ to $\approx 100K$.

\section{Sanity check: augmentations improve accuracy}
\label{a:sanity}

\begin{figure}[tbp]
\begin{subfigure}{0.3\textwidth}
\centering
\includegraphics[width=1.8in]{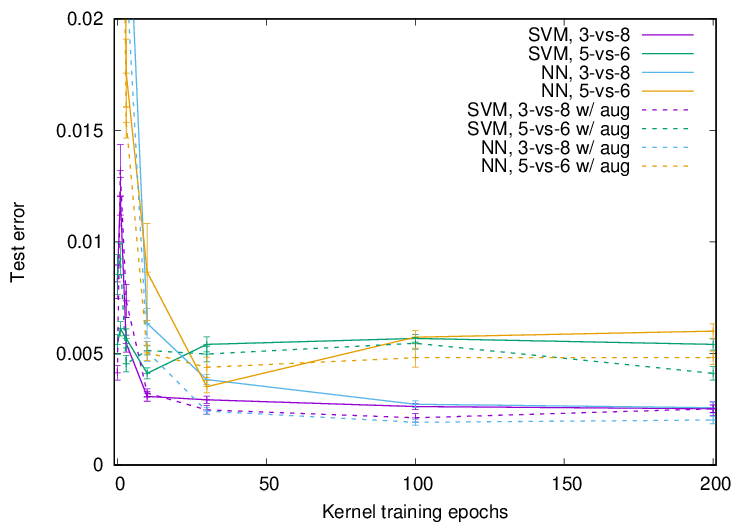}
\caption{A plot of the test errors of the SVM model, and the NN model that
trained the kernel for the SVM model, with and without data augmentation,
on the VGG-like models.}
\label{f:kernel_epochs_vs_aug_noaug_accuracies_vgg}
\end{subfigure}
\begin{subfigure}{0.3\textwidth}
\centering
\includegraphics[width=1.8in]{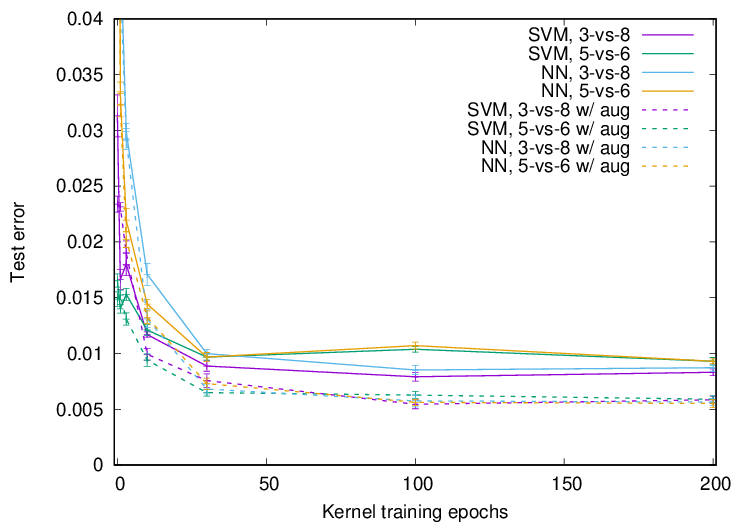}
\caption{A plot of the test errors of the SVM model, and the NN model that
trained the kernel for the SVM model, with and without data augmentation,
on the fully connected model.}
\label{f:kernel_epochs_vs_aug_noaug_accuracies_mlp}
\end{subfigure}
\begin{subfigure}{0.3\textwidth}
\centering
\includegraphics[width=1.8in]{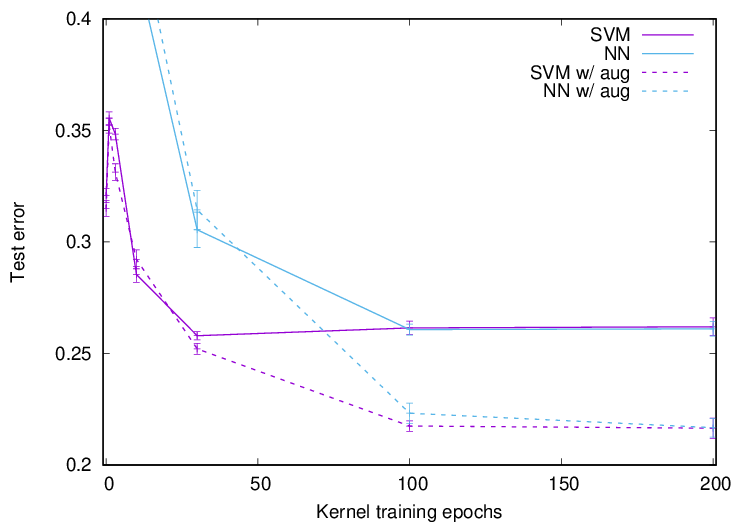}
\caption{A plot of the test errors of the SVM model, and the NN model that
trained the kernel for the SVM model, with and without data augmentation,
for the VGG-like model trained on cats-vs-dogs.}
\label{f:kernel_epochs_vs_aug_noaug_accuracies_cifar10}
\end{subfigure}
\caption{Test error of models with and without data augmentation.}
\label{f:kernel_epochs_vs_aug_noaug_accuracies}
\end{figure}

Figure~\ref{f:kernel_epochs_vs_aug_noaug_accuracies} reproduces the well-known
fact that data augmentations improve the test accuracy of learned models.

\end{document}